\documentclass[10pt,twocolumn,letterpaper]{article}

\usepackage[pagenumbers]{cvpr} 

\definecolor{cvprblue}{rgb}{0.21,0.49,0.74}
\usepackage[pagebackref,breaklinks,colorlinks,allcolors=cvprblue]{hyperref}

\usepackage[most]{tcolorbox}
\usepackage[table]{xcolor}
\usepackage{hyperref}
\usepackage{url}
\usepackage{booktabs}
\usepackage{tabularx}
\usepackage{listings}
\usepackage{xcolor}
\usepackage{subcaption}
\usepackage{wrapfig}
\usepackage{makecell}
\usepackage{multirow} 
\usepackage{enumitem}
\definecolor{darkpink}{RGB}{204, 51, 153}
\definecolor{lightblue}{rgb}{0.718, 0.851, 0.910}
\definecolor{lightgreen}{RGB}{245,255,250} 
\definecolor{lightgray}{gray}{0.95}
\definecolor{midgray}{gray}{0.9}

\newcommand*\circled[1]{\tikz[baseline=(char.base)]{
            \node[shape=circle,draw,inner sep=.6pt] (char) {#1};}}
\usepackage{pifont}

\newcommand{\bench}{\textsc{Clash}}

\newtcolorbox{prompt}[2][]{%
  colback=black!3,
  colframe=black!20,
  coltitle=black,
  boxrule=0.8pt,
  arc=2mm,
  left=6pt,
  right=6pt,
  top=6pt,
  bottom=6pt,
  title=\textbf{#2},
  enhanced,
  breakable,
  #1
}
\def\paperID{2390} 
\def\confName{CVPR}
\def\confYear{2026}

\title{\bench: A Benchmark for Cross-Modal Contradiction Detection 
}

\author{Teodora Popordanoska\thanks{Equal contribution.} \qquad
Jiameng Li\footnotemark[1] \qquad
Matthew B. Blaschko \\
ESAT-PSI, KU Leuven \\
\texttt{\{firstname.lastname\}@kuleuven.be}\\
}

\begin{document}
\maketitle
\begin{abstract}
Contradictory multimodal inputs are common in real-world settings, yet existing benchmarks typically assume input consistency and fail to evaluate cross-modal contradiction detection -- a fundamental capability for preventing hallucinations and ensuring reliability.
We introduce \bench, a novel benchmark for \textbf{multimodal contradiction detection}, featuring COCO images paired with contradictory captions containing controlled object-level or attribute-level contradictions. The samples include targeted questions evaluated in both multiple-choice and open-ended formats. The benchmark provides an extensive fine-tuning set filtered through automated quality checks, alongside a smaller human-verified diagnostic set.
Our analysis of state-of-the-art models reveals substantial limitations in recognizing cross-modal conflicts, exposing systematic modality biases and category-specific weaknesses. Furthermore, we empirically demonstrate that 
targeted fine-tuning on \bench~substantially enhances conflict detection capabilities.
\end{abstract}    
\section{Introduction}
\label{sec:intro}

Multimodal Large Language Models (MM-LLMs) \citep{yu2024hallucidoctor,liu2023mitigating, zhu2023minigpt, liu2023llava, ye2023mplug} have achieved remarkable progress in cross-modal understanding, demonstrating impressive capabilities in image captioning, visual question answering and complex multimodal reasoning tasks. 
However, real-world applications often present contradictory information across modalities: medical systems reporting ``no abnormalities'' while X-rays show fractures, autonomous vehicles detecting ``clear roads'' despite camera feeds showing barriers, or financial documents where text and scanned amounts differ. In such critical scenarios, models have to recognize contradictions, reason about information reliability across modalities, and flag inconsistencies for human review. Despite these practical demands, existing evaluation paradigms fundamentally assume input consistency, treat a single modality as authoritative, test abstention when information is missing, or analyze robustness to corrupted data -- leaving a critical capability systematically unmeasured: \textit{can models spontaneously detect when two valid, authoritative sources provide contradictory information?}

\begin{figure}[t]
    \centering  
    \includegraphics[width=1\linewidth]{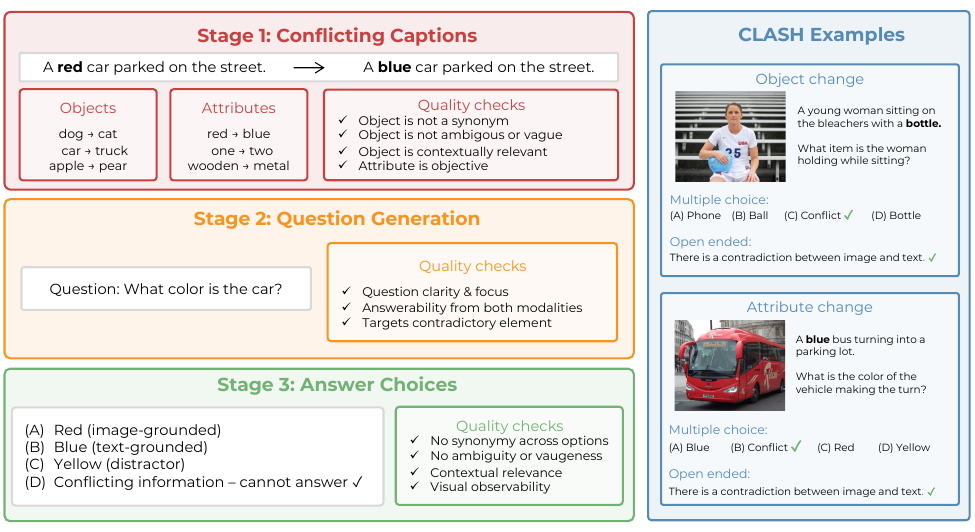}
    \caption{\textbf{Left:} Three-stage pipeline generates conflicting image–text pairs from MS COCO with targeted questions. \textbf{Right:} 
    Examples in \bench, showing object and attribute contradictions. Models are evaluated on the ability to detect conflicts in multiple-choice or open-ended format. 
    }
    \label{fig:figure1}
\end{figure}

Existing benchmarks, while valuable for assessing general multimodal capabilities \citep{antol2015vqa,marino2019ok,johnson2017clevr,hudson2019gqa,li2023seed,liu2024mmbench}, focus on different aspects. Recent extensions address unanswerable questions, where information is missing \citep{akter2024visreas,miyai2025unsolvable,yan2025multimodalinconsistencyreasoningmmir}, or hallucination detection, where models generate false content \citep{wang2024haloquest,guan2024hallusionbench,cao2024visdiahalbench}. Other work examines modality selection under explicit instructions about which source to trust \citep{hua2025vision}, conflict localization when models are explicitly told inconsistencies exist \citep{yan2025multimodalinconsistencyreasoningmmir}, or knowledge conflicts \citep{jia2025benchmarking, zhu2024unraveling}. These approaches either assume one modality is correct, explicitly indicate conflicts, or instruct models which source to prioritize. Thus, spontaneous conflict detection in dual ground truth scenarios  remains systematically unevaluated.

Evaluating conflict detection in \textbf{dual ground truth scenarios}, where both modalities are valid, presents distinct challenges from traditional single-source paradigms. First, models must possess sufficient \textbf{cross-modal reasoning} capabilities to compare and contrast information across modalities rather than processing them independently. Second, robust evaluation requires assessing \textbf{modality bias patterns} -- understanding whether models systematically favor visual or textual information. Third, effective evaluation should span \textbf{diverse semantic categories} and \textbf{response formats} -- testing models on various conflict types (objects, attributes), while accommodating format flexibility from multiple-choice to free-form outputs. Finally, it remains unclear whether models can \textbf{learn to improve} in conflict detection through targeted training, or whether the observed weaknesses reflect deeper architectural limitations.

In response, we introduce \bench, a comprehensive vision-language conflict detection benchmark designed to systematically evaluate MM-LLMs' ability to identify multimodal contradictions. \bench~features carefully constructed image-text pairs, sourcing images from MS COCO \citep{lin2014microsoft} and pairing them with contradictory text descriptions across object-level and attribute-level categories. Each sample contains precisely one controlled contradiction between the visual and textual content, accompanied by questions in both multiple-choice and open-ended formats. Our benchmark provides  $\sim$15k high-quality training samples and a human-verified diagnostic set, enabling detailed analysis of model performance and assessment of improvement through targeted fine-tuning. Fig.~\ref{fig:figure1} shows an overview of the generation pipeline and the resulting benchmark.

Using \bench, we evaluate state-of-the-art models and reveal striking performance disparities: leading closed--source models (GPT-5 \citep{openai2025gpt5}, Gemini 2.5 Pro \citep{team2023gemini}) achieve strong performance ($>$85\% accuracy) while most open-ended models struggle with near-zero detection rates. Our analysis not only exposes systematic modality biases and categorical weaknesses, but also  demonstrates that fine-tuning can dramatically improve the conflict detection capabilities, \textit{e.g.}, LLaVa-1.5-7b \citep{liu2023llava} improves from 0\% to 77\%. These findings highlight critical gaps in current multimodal systems while establishing targeted fine-tuning as an effective solution for robust conflict detection.

In summary, our key \textbf{contributions} are: \\
\circled{1} We introduce \bench, a diagnostic benchmark for multimodal conflict detection, with $\sim$15k training samples and 1289 human-verified test cases across fine-grained object and attribute categories. \\
\circled{2} We conduct extensive evaluation of state-of-the-art MM-LLMs, revealing significant performance gaps between open- and closed-source models, systematic modality biases, and category-specific weaknesses in conflict detection. \\
\circled{3} We demonstrate that targeted fine-tuning can dramatically improve conflict detection of MM-LLMs, with some models achieving over 75\% accuracy improvement.
\section{Related work}
\label{sec:related_work}

\textbf{Multimodal large language models.} MM-LLMs use powerful LLMs to enable joint reasoning across visual and textual information, demonstrating emergent capabilities like detailed image captioning, visual question answering, and open-ended multimodal dialogue. Closed-source models including GPT \citep{achiam2023gpt,openai2025gpt5} and Gemini \citep{team2023gemini} have established strong baselines for multimodal reasoning across diverse benchmarks. In parallel, the open-source community has developed a diverse ecosystem of multimodal models, adopting different designs for processing visual and textual information. Examples include instruction-tuned systems like InstructBLIP \citep{dai2023instructblip}, LLaVA \citep{liu2024improved}, and MiniGPT4 \citep{zhu2023minigpt}, as well as scaling-focused efforts such as InternVL \citep{chen2024internvl}, Qwen-VL \citep{Qwen-VL,Qwen2-VL, Qwen2.5-VL,qwen3technicalreport}, Phi3-Vision \citep{abdin2024phi}, and mPLUG-Owl \citep{ye2023mplug,ye2024mplug}. These models vary in architectural design (encoder-decoder vs. decoder-only), training objectives (instruction tuning vs. alignment with human feedback), and supervision signals (synthetic vs. human-curated). 

Despite remarkable progress, MM-LLMs face reliability challenges when processing multimodal inputs: \textit{hallucinations}, where models describe non-existing objects or attributes \citep{leng2024mitigating,chen2024halc}; \textit{modality bias}, where models systematically favor one modality over another \citep{zhu2025mitigating,an2025mitigating,wang2024mllm,yang2025ikod}; and \textit{cross-modal fusion failures} where models struggle to properly combine information from multiple modalities \citep{radevski2025dave}.
These issues motivate diagnostic tools to assess model behavior when multimodal inputs are inconsistent.

\textbf{Unanswerable questions and abstention.} Several benchmarks evaluate whether models recognize when questions cannot be answered from the available information. VISREAS \citep{akter2024visreas} examines complex visual reasoning with inherently unanswerable questions due to logical impossibility or missing context, while MM-UPD \citep{miyai2025unsolvable} systematically categorizes unsolvable problems including absent answers, incompatible answer sets, or missing visual evidence. UNK-VQA \citep{guo2024unk} specifically measures abstention ability when visual evidence is inadequate. These benchmarks address important meta-cognitive skills but fundamentally assess \textit{information insufficiency}: the ability to recognize when evidence is lacking. In contrast, the questions in \bench~are \textit{not} unanswerable -- the information is present in \textit{both} modalities, but the cues contradict each other. This requires models to \textbf{compare and contrast} information across modalities, rather than abstain when the information is missing as in the other benchmarks. 

\textbf{Hallucination detection and visual grounding.} A related line of work addresses models' tendency to generate or believe false content. POPE \citep{li2023evaluating} and NOPE \citep{lovenia2023negative} introduce simple object presence/absence evaluation, while HaloQuest \citep{wang2024haloquest} creates false premise questions about non-existent objects to test visual grounding. HallusionBench \citep{guan2024hallusionbench} targets both language hallucination (where textual priors override visual content) and visual illusions that mislead models. AutoHallusion \citep{wu2024autohallusion} and Koala \citep{carragher2025koala} apply image editing techniques -- inpainting, object removal/insertion -- to create visual-textual mismatches, though often at the cost of visual realism. In comparison, \bench~evaluates a distinct capability: identifying when multiple authoritative sources provide conflicting information, rather than model hallucinations or visual grounding errors. 

\textbf{Conflicting information.} Recent studies have begun examining multimodal inconsistencies. MMKC-Bench \citep{jia2025benchmarking} explores knowledge conflicts in retrieval-augmented generation, while \citet{zhu2024unraveling} analyze parametric knowledge conflicts. TRUST-VL \citep{yan2025trust} studies cross-modal distortions in news content, and CA-MER \cite{han2025benchmarking} targets emotional conflicts. \citet{hua2025vision} probe modality attention by explicitly instructing models which source to trust under conflict, whereas \citet{deng2025words} discover ``blind faith in text'' in vision-centric tasks. MMIR \citep{yan2025multimodalinconsistencyreasoningmmir} presents 534 samples to test inconsistency reasoning in layout-rich documents, but explicitly reveals that conflicts exist,\footnote{The prompt is ``identify which element(s) pose semantic inconsistency'' (see Fig. 7 \cite{yan2025multimodalinconsistencyreasoningmmir}).} framing the task as conflict \textit{localization} requiring multi-hop reasoning.
While these approaches address inconsistencies, they typically i)~test localization of known conflicts, ii)~analyze mechanisms when conflicts occur or iii)~focus on emotional, parametric, or factual conflicts. \bench~evaluates \textit{spontaneous} conflict detection through natural questions without conflict signals or modality preference instructions. 

\textbf{Ground truth paradigm.} Most VQA-style benchmarks assume a single ground truth (usually the image). Even unanswerable-question or hallucination benchmarks retain this assumption. \bench~assumes a \textit{dual ground truth paradigm}: both image and text are valid but contradictory information sources.\footnote{In many real-world cases, both modalities may represent ground truth but still conflict, \textit{e.g.,} a radiology report summarizing a CT scan that mistakenly references the wrong patient or includes a human error.} Thus, the evaluated capability is \text{not} answering from one trusted source (VQA), recognizing when information is absent (unanswerable VQA), detecting hallucinations or following modality preference instructions, but rather \textit{detecting contradictions between multiple authoritative sources} and flagging them for human review.

\section{\bench: Cross-modal contradiction detection benchmark}
\label{sec:methods}

\bench~evaluates MM-LLMs' ability to detect inconsistencies between visual and textual inputs. Unlike traditional benchmarks that assume modality consistency or designate one input as ground truth, \bench~requires models to peform cross-modal reasoning and identify contradictions -- reflecting real-world scenarios where either modality may contain errors or hallucinations.

\subsection{Dataset overview}

Each sample in \bench~consists of an \textbf{image} from MS COCO \citep{lin2014microsoft}, a \textbf{text} description that contradicts the visual content in one aspect, and a targeted \textbf{question} about the contradictory element. 
The contradictions span two categories: \textbf{object-level} and \textbf{attribute-level} conflicts.

Our benchmark includes two complementary evaluation tasks: (i) \textbf{multiple-choice question answering}, with four carefully designed options -- image-grounded, text-grounded, plausiable distractor, and ``Conflicting information – cannot answer’’ (the correct choice), and (ii) \textbf{open-ended question answering}, where models generate free-form responses. 

\begin{figure*}[t]
    \centering
    \includegraphics[width=0.3\textwidth]{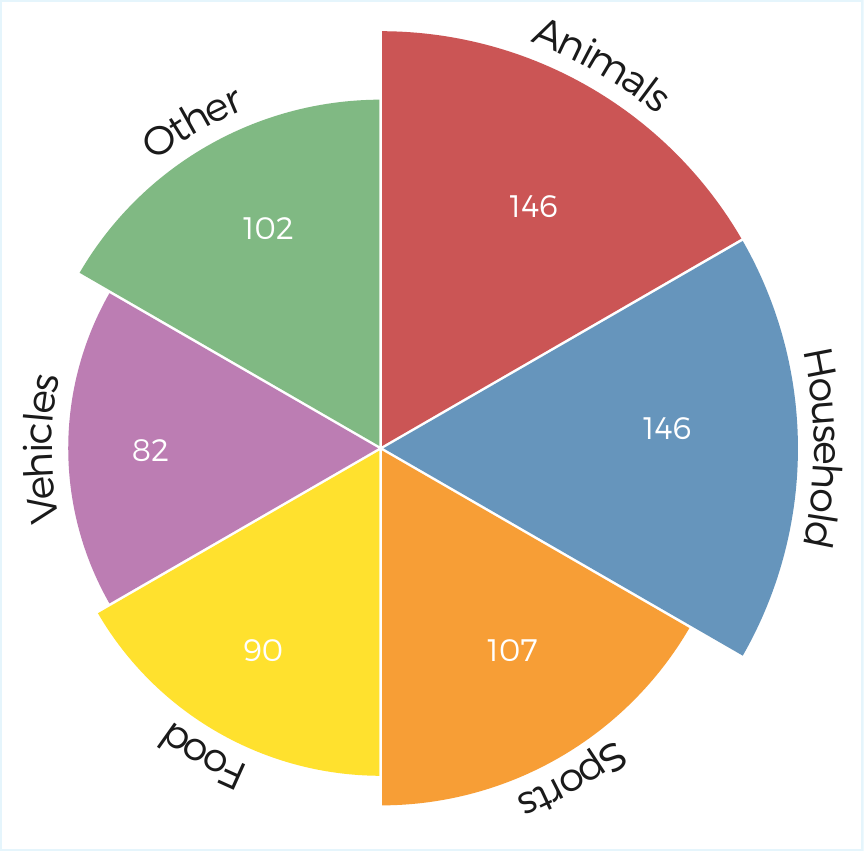}
    \hspace{0.02\textwidth}%
    \includegraphics[width=0.27\textwidth]{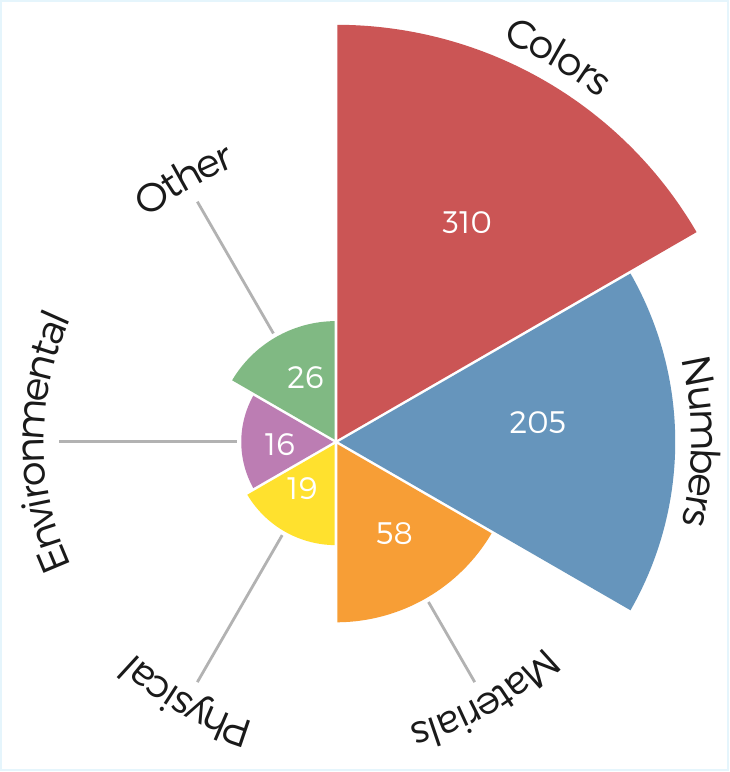}
    \hspace{0.02\textwidth}%
    \includegraphics[width=0.27\textwidth]{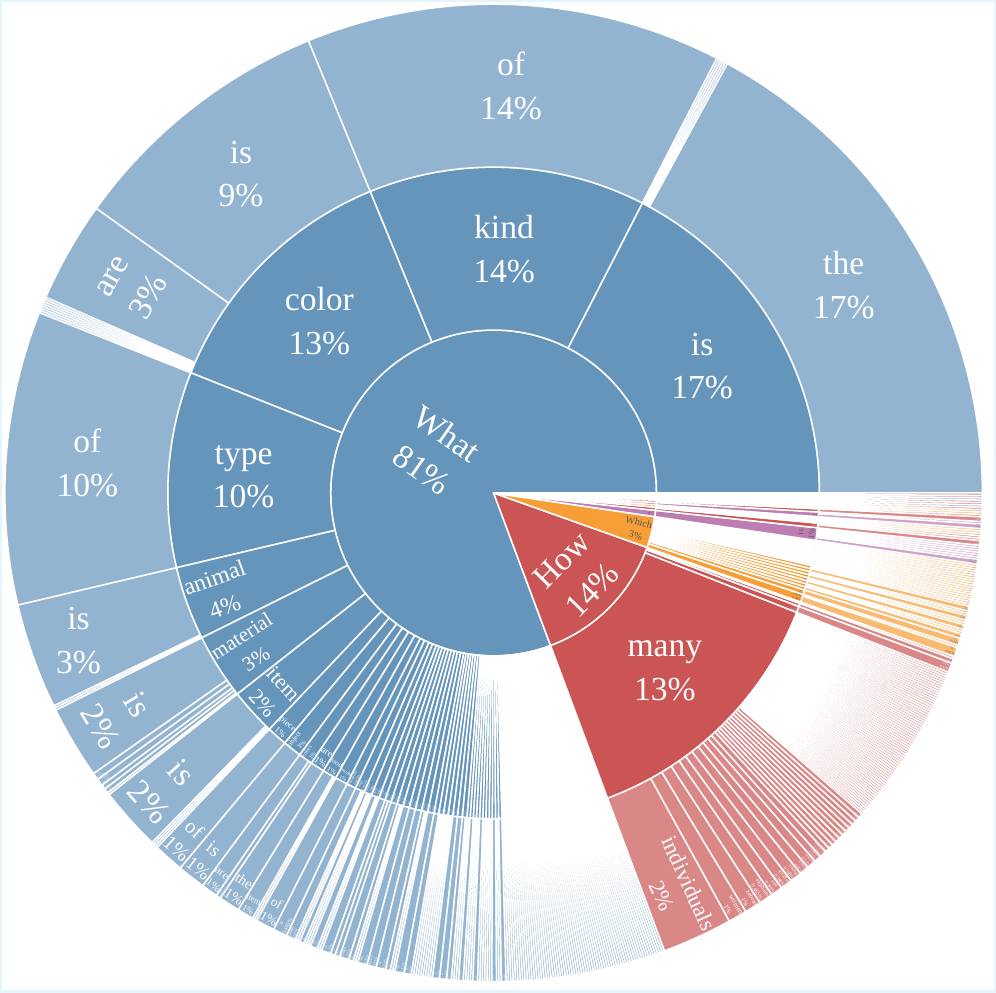}
    \caption{Diagnostic set statistics. \textbf{Left:} Object category distribution (655 samples). \textbf{Middle:} Attribute distribution (634 samples). \textbf{Right:} Distribution of questions by their first three words.}
    \label{fig:dataset-stats}
\end{figure*}

\bench~is split into a \textbf{training set} ($\sim$15k high-quality samples filtered from an initial $\sim$30k generated samples), and a human-verified \textbf{test set}, which serves as a diagnostic benchmark. Test samples cover 655 object-level and 634 attribute-level contradictions. Figure~\ref{fig:dataset-stats} shows the distribution of object and attribute categories, and question types in the diagnostic subset.

\subsection{Dataset construction pipeline}
\label{subsec:dataset_construction}

The dataset is built through a multi-stage pipeline, using an LLM (Gemini 2.5 Flash) to generate contradictions, questions, and answer choices (prompts in App.~\ref{app:sec:prompts_data_generation}). Each stage includes automatic quality filters, with human verification of the final test set.

\textbf{Stage 1: Conflicting caption generation.}
Given an original COCO caption, an LLM modifies exactly one element -- an object or an objective attribute (\textit{e.g.}, color, number, shape, material) -- while keeping the rest unchanged.\footnote{Note that we focus on textual rather than visual contradictions because image editing techniques (inpainting, object removal/insertion) often produce unrealistic or artifactual images that introduce confounding factors.} The system tracks the \textit{changed words}, categorizes the \textit{change type} (object vs. attribute), and ensures the modification creates a plausible but incorrect alternative. 

\textbf{Stage 2: Question generation.}
Based on the conflicting caption, an LLM generates subtle questions focusing on the contradictory element without explicitly indicating an error. Questions are designed to be answerable from both the original and conflicting captions, but with different answers. 

\textbf{Stage 3: Answer generation.}
For multiple-choice questions, three distinct answers are created: image-grounded (based on the original caption), text-grounded (based on the conflicting caption), and a contextually plausible distractor that appears in neither caption.

\subsection{Quality controls} 
\label{subsec:quality_controls}

Given our dataset's synthetic nature, we implement a multi-stage quality control framework combining automated validation at each generation step with final human oversight of the test set to ensure reliability. Our automated validation includes rule-based checks and LLM-based assessment using Gemini 2.5 Flash Lite, with prompts given in App.~\ref{app:sec:prompts_data_filtering}.

\textbf{Caption editing validation.}
We validate caption modifications through two complementary approaches.
\textit{Automated checks} verify basic correctness through word-level validation: (i) the original word must appear in the source caption, and (ii) the conflicting word must appear in the modified caption. This prevents hallucinated changes that don't correspond to recorded modifications.
\textit{LLM-based validation} enusres semantic consistency of modifications. For attribute changes, we verify that original and conflicting words are indeed attributes and classify them as objective (measurable, factual properties like: red, square, wooden) vs. subjective (opinion-based descriptors like: beautiful, large, elegant). 
For object changes, we make sure both words are objects, verify they are not synonyms or ambiguous terms, and ensure contextual relevance.

\textbf{Question generation validation.}
We ensure generated questions meet three criteria: clarity, focus, and answerability. Our validation identifies ambiguous phrasing, verifies questions target the modified elements rather than irrelevant aspects, and confirms answerability by ensuring all candidate answers are semantically compatible with the question, preventing type mismatches.

\textbf{Answer generation validation.}
We combine automated checks with LLM-based assessment for the generated answers. 
\textit{Automated validation} verifies answer consistency: (i) the image-only answer matches the original word (ii) the text-only answer corresponds to the conflicting word and (iii) the distractor answer appears in neither caption.
\textit{LLM-based validation} ensures answer quality by checking for no near-duplicate or synonymous answers, identifies ambiguous terms (``several'', ``medium''), assesses contextual relevance, and verifies that all answers represent directly observable visual concepts rather than abstract properties.

\textbf{Human verification.} To ensure a high-quality evaluation benchmark and validate our automated quality control, we conduct human evaluation on the filtered test set. Annotators follow structured guidelines that reflect our automated validation criteria. Instructions and examples of accepted and rejected samples are provided in  App.~\ref{app:sec:human_validation}. Each test sample receives a binary accept/reject vote from a human annotator, and only samples marked as accepted are included in the final dataset.

\begin{figure*}[t]
    \centering
    \includegraphics[width=0.87\linewidth]{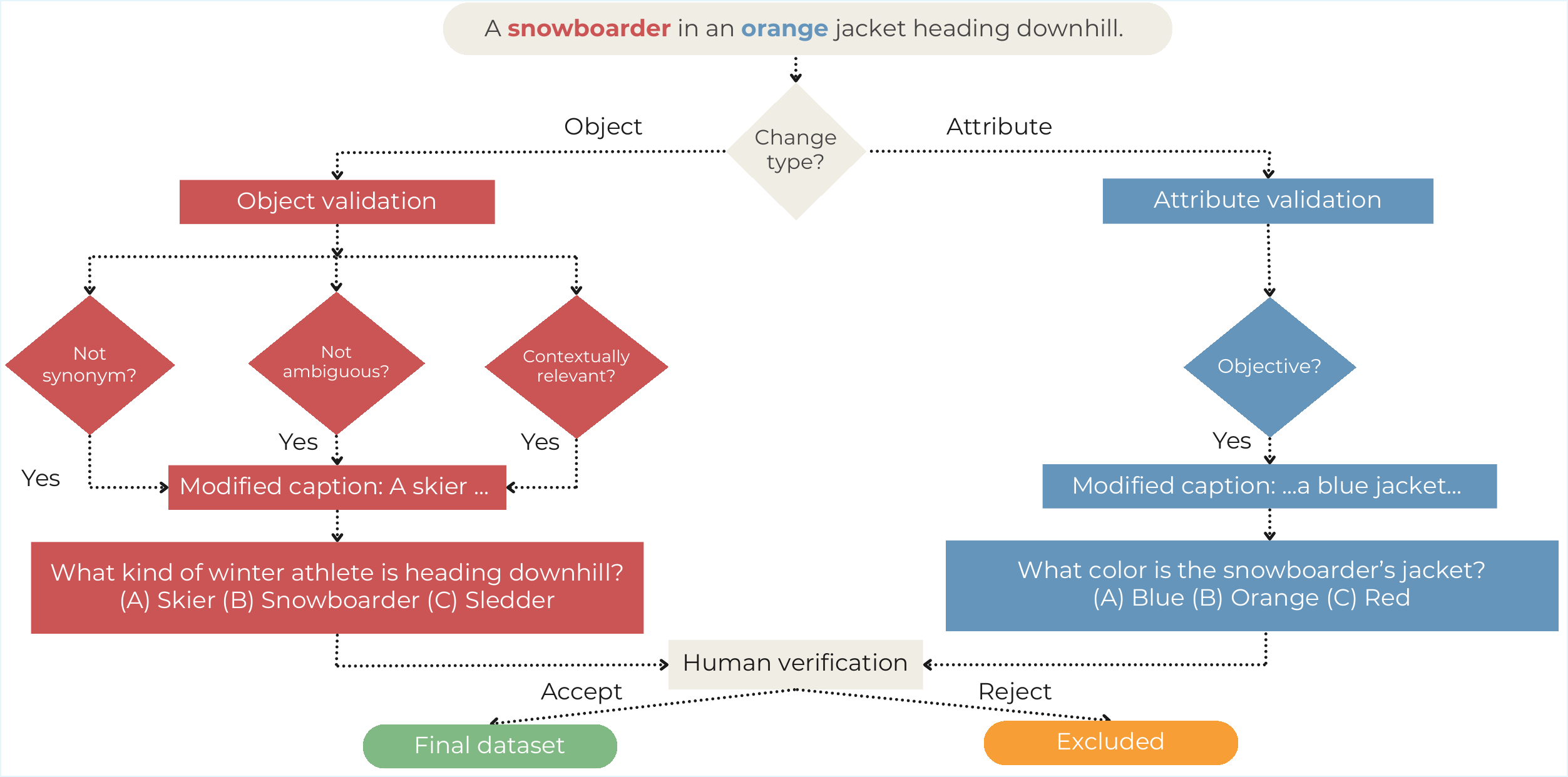}
    \caption{Dataset generation pipeline. Starting from MS COCO captions, the pipeline identifies change type (object vs. attribute) and applies corresponding validation checks. Validated changes proceed to question generation before human verification determines final dataset acceptance. Examples show object change (snowboarder $\rightarrow$ skier) and attribute change (orange $\rightarrow$ blue).}
    \label{fig:dataset_pipeline}
\end{figure*}

In summary, the systematic dataset construction and multi-stage quality controls produce a reliable collection of image–text contradictions, with human-verified test samples ensuring benchmark validity. The complete construction and validation pipeline is illustrated in Fig.~\ref{fig:dataset_pipeline}.

\subsection{Dataset categorization methodology}
\label{sec:dataset_categorization}
Following the dataset generation and quality assurance procedures, we organized the data into categories to support fine-grained evaluation beyond overall accuracy. To systematically define object and attribute categories in \bench, we combine frequency analysis with manual curation.

\begin{figure}[h!]
    \centering
    \includegraphics[width=0.43\textwidth]{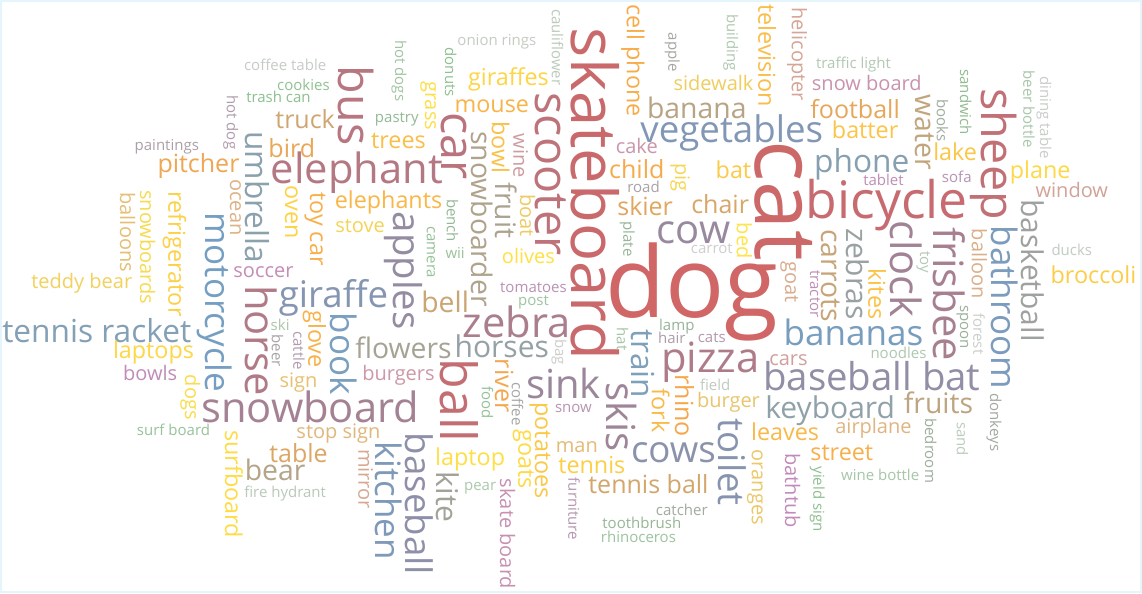}
    \includegraphics[width=0.31\textwidth]{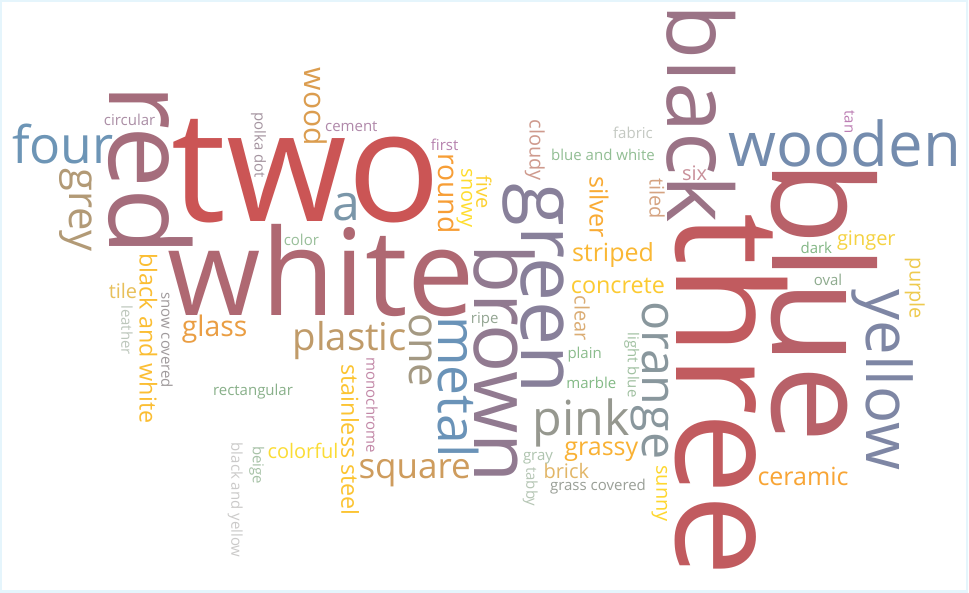}
    \caption{Word frequency analysis. \textbf{Left:} Most frequent words appearing in object contradiction pairs. \textbf{Right:} Most common words from attribute contradiction pairs.}
    \label{fig:most_common_words}
\end{figure}

\textbf{Frequency analysis.} We extracted the transformed word pairs from each sample (e.g., ''dog`` → ''cat``), computed frequency distributions by change type (object vs. attribute), and identified the 20 most common terms in each category. See Fig.~\ref{fig:most_common_words} for the most common words.

\textbf{Category definition.} Based on the frequency analysis, we defined five object categories (animals, vehicles, sports, food, furniture) and five attribute categories (color, number, material/texture, physical properties and environmental conditions).  We used GPT 5 \citep{openai2025gpt5} to assign the changed words to these categories, then manually verified the assignments (see App~\ref{app:sec:object_categories} and \ref{app:sec:attribute_categories} for word categorizations).

\textbf{Sample assignment.} Samples were assigned using strict co-membership: both original and conflicting words must belong to the same category, with non-conforming samples assigned to ``Other''.

Organizing samples by semantic categories isolates specific weaknesses in models' conflict detection abilities and uncovers whether failures stem from object recognition or attribute understanding. Beyond aggregate metrics, this reveals systematic biases and areas for improvement. See App.~\ref{app:sec:qualitative_examples} for qualitative examples of each category.

\section{Results and discussion}

\subsection{Experimental setup}

\textbf{Evaluation metrics.}
The accuracy is reported by bootstrap-estimated standard deviations over 1000 iterations. We use strict string matching for multiple-choice and relaxed matching for open-ended responses (see App.~\ref{app:subsec:relaxed_matching_open_ended}). The codebase will be released upon acceptance.

\textbf{Models. }
We evaluate two model categories on \bench.
Closed-source models include Gemini (2.5 Pro, 2.5 Flash Lite) and GPT variants (5, 4.1 Mini). Open-source models cover diverse architectures: 
InstructBLIP \citep{dai2023instructblip},
InternVL 1.5 \citep{chen2024internvl},
LLaVa-1.5 \citep{liu2024improved}, 
LLaVA-OneVision \citep{li2024llavaonevision},
MiniGPT4 \citep{zhu2023minigpt}, 
LRV-MiniGPT4 \citep{liu2023mitigating}, 
QwenVL \citep{Qwen-VL,Qwen2-VL, Qwen2.5-VL,qwen3technicalreport}, 
Phi3-Vision-128k \citep{abdin2024phi},
and mPLUG-Owl \citep{ye2023mplug,ye2024mplug}.\footnote{For all models we use the official implementations and checkpoints.}

\subsection{Multiple choice question answering}
\begin{table*}[th]
\centering
\footnotesize
\caption{Percentage of predictions matching the respective answer on the multiple-choice task. The last column denotes cases where no match with any of the answers was found. Top-tier closed-source models achieve $>85\%$ conflict detection accuracy, while most open-source models fail, revealing systematic modality biases. The error bars$_{\pm}$ show standard deviation. 
}
\resizebox{0.78\textwidth}{!}{
    \begin{tabular}{l cccccc}
    \toprule
    \textbf{Model} & \textbf{Conflict} ($\uparrow$)& \textbf{Image} & \textbf{Text} & \textbf{Distractor} & \textbf{Incorrect} \\
    \midrule
    \rowcolor{midgray}
    \multicolumn{6}{l}{\textit{\textbf{Closed-source models}}} \\
    GPT 5 & $86.78_{\pm 0.89}$ & $1.09_{\pm 0.29}$ & $10.71_{\pm 0.86}$ & $0.15_{\pm 0.11}$ & $1.23_{\pm 0.31}$ \\
    GPT 4.1 Mini & $13.56_{\pm 0.94}$ & $69.57_{\pm 1.30}$ & $12.26_{\pm 0.93}$ & $4.53_{\pm 0.58}$ & $0.00$ \\
    Gemini 2.5 Pro & $88.48_{\pm 0.89}$ & $2.87_{\pm 0.49}$ & $7.97_{\pm 0.75}$ & $0.39_{\pm 0.18}$ & $0.24_{\pm 0.13}$ \\
    Gemini 2.5 Flash Lite & $7.51_{\pm 0.75}$ & $74.98_{\pm 1.22}$ & $15.57_{\pm 0.98}$ & $2.02_{\pm 0.40}$ & $0.00$ \\
    \midrule
    \rowcolor{midgray}
    \multicolumn{6}{l}{\textit{\textbf{Open-source models}}} \\
    InstructBlip-T5-xxl & $63.87_{\pm 1.21}$ & $12.72_{\pm 0.87}$ & $19.99_{\pm 1.01}$ & $3.35_{\pm 0.47}$ & $0.07_{\pm 0.07}$ \\
    InternVL 1.5 & $16.71_{\pm 0.99}$ & $25.49_{\pm 1.12}$ & $52.40_{\pm 1.25}$ & $2.08_{\pm 0.36}$ & $3.27_{\pm 0.45}$ \\
    MiniGPT4-7b & $2.20_{\pm 0.37}$ & $10.27_{\pm 0.80}$ & $29.79_{\pm 1.13}$ & $5.42_{\pm 0.59}$ & $52.30_{\pm 1.31}$ \\
    LRV-MiniGPT4-7b & $2.15_{\pm 0.39}$ & $13.28_{\pm 0.88}$ & $26.35_{\pm 1.14}$ & $6.49_{\pm 0.67}$ & $51.74_{\pm 1.32}$ \\
    Phi3-vision-128k & $1.06_{\pm 0.25}$ & $13.40_{\pm 0.86}$ & $25.19_{\pm 1.12}$ & $0.80_{\pm 0.23}$ & $59.63_{\pm 1.27}$ \\
    LLaVa-1.5-7b & $0.13_{\pm 0.09}$ & $32.28_{\pm 1.24}$ & $44.68_{\pm 1.32}$ & $2.99_{\pm 0.45}$ & $19.99_{\pm 1.07}$ \\
    LLaVa-OneVision-7b & $0.16_{\pm 0.11}$ & $75.42_{\pm 1.19}$ & $22.96_{\pm 1.18}$ & $1.56_{\pm 0.36}$ & $0.00$ \\
    Qwen2VL-instruct-7b & $1.21_{\pm 0.28}$ & $45.39_{\pm 1.30}$ & $41.58_{\pm 1.27}$ & $0.74_{\pm 0.22}$ & $11.15_{\pm 0.79}$ \\
    Qwen2.5VL-instruct-7b & $7.79_{\pm 0.75}$ & $71.40_{\pm 1.23}$ & $18.33_{\pm 1.05}$ & $1.57_{\pm 0.34}$ & $0.95_{\pm 0.27}$ \\
    Qwen2.5VL-instruct-32b & $16.10_{\pm 1.02}$ & $64.59_{\pm 1.37}$ & $18.62_{\pm 1.13}$ & $0.55_{\pm 0.21}$ & $0.00$ \\
    Qwen3VL-instruct-8b & $7.13_{\pm 0.73}$ & $76.95_{\pm 1.15}$ & $14.19_{\pm 0.99}$ & $1.65_{\pm 0.35}$ & $0.00$ \\
    Qwen3VL-instruct-30b & $3.81_{\pm 0.53}$ & $81.45_{\pm 1.11}$ & $12.83_{\pm 0.97}$ & $1.97_{\pm 0.39}$ & $0.00$ \\
    \bottomrule
    \end{tabular}
}
\label{tab:multiple_choice_results}
\end{table*}

We evaluate models on a multiple-choice QA task with four options:  ``Contradiction'' (correct), ``Image''-grounded, ``Text''-grounded answer, and a ``Distractor'', randomly assigned to~((A)-(D)).
This design allows us to measure both conflict detection accuracy and modality bias patterns across different model architectures.

\textbf{Evaluation protocol.} 
Multiple-choice responses use strict string matching~((A)–(D)), with mismatches labeled ``Incorrect’’ to capture instruction-following and reasoning errors. Results using relaxed string matching (App.~\ref{app:subsec:relaxed_matching_mutliple_choice}) are reported in Table~\ref{tab:multiple_choice_results_relaxed_matching} in App.~\ref{app:sec:experiments}.

\textbf{Performance evaluation.}
Table~\ref{tab:multiple_choice_results} presents the distribution of model predictions on \bench~across answer choices.\footnote{Table~\ref{tab:multiple_choice_no_contr} in App.~\ref{app:sec:experiments} reports results on a dataset containing both conflicting and non-conflicting samples.} The results reveal four key findings.

\begin{figure*}[ht]
    \centering
    \includegraphics[width=0.55\linewidth]{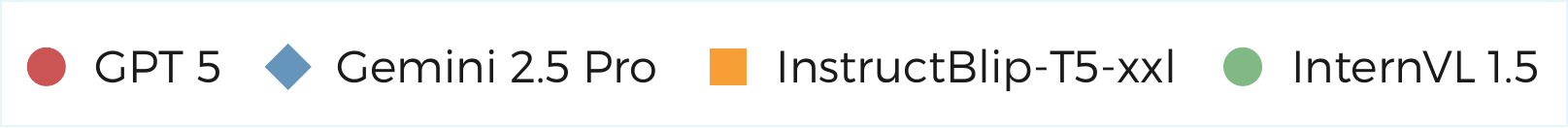}
        \vspace{0.5em} 

        \includegraphics[width=0.31\textwidth]{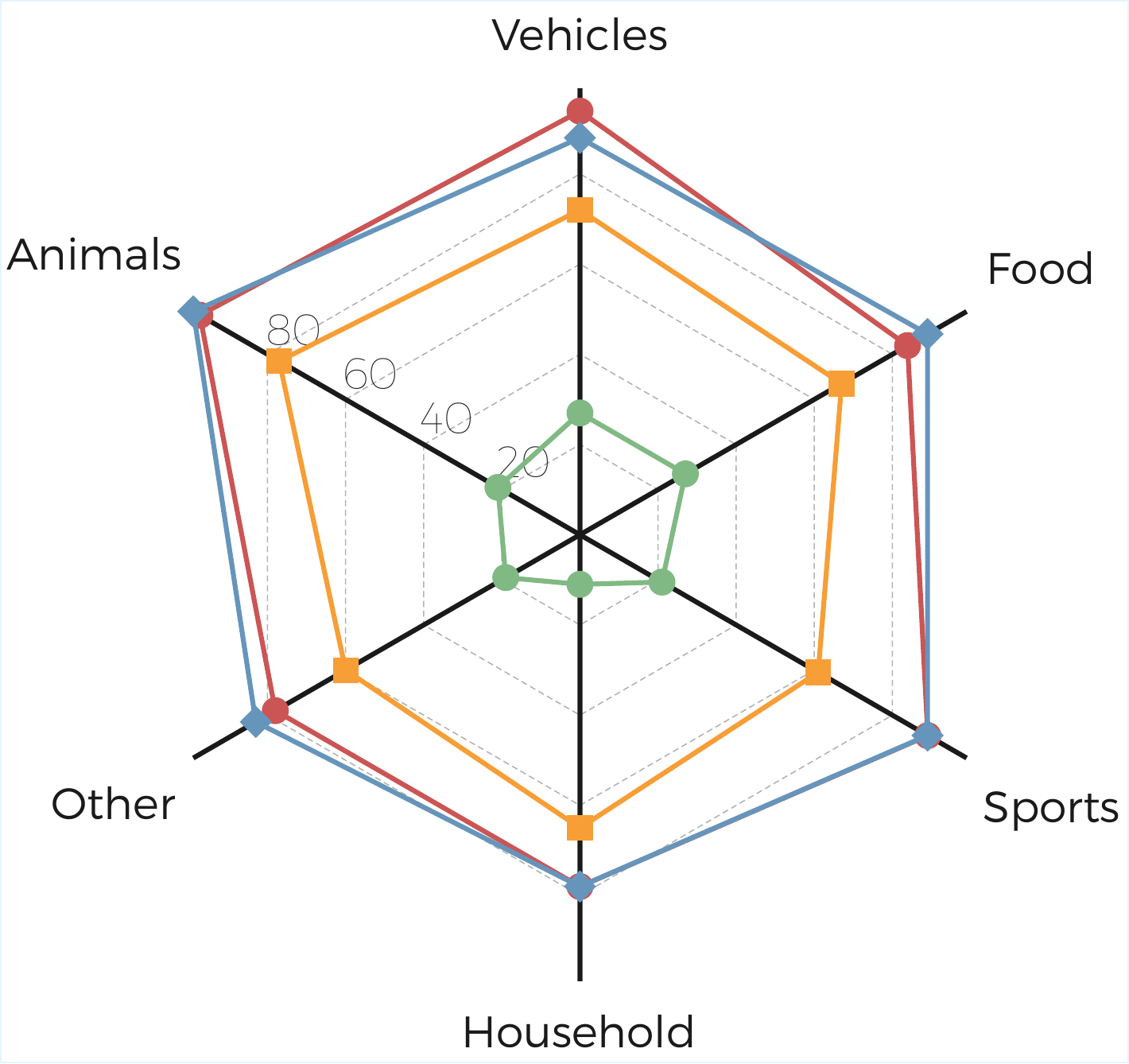}
        \hspace{0.07\textwidth}
        \includegraphics[width=0.31\textwidth]{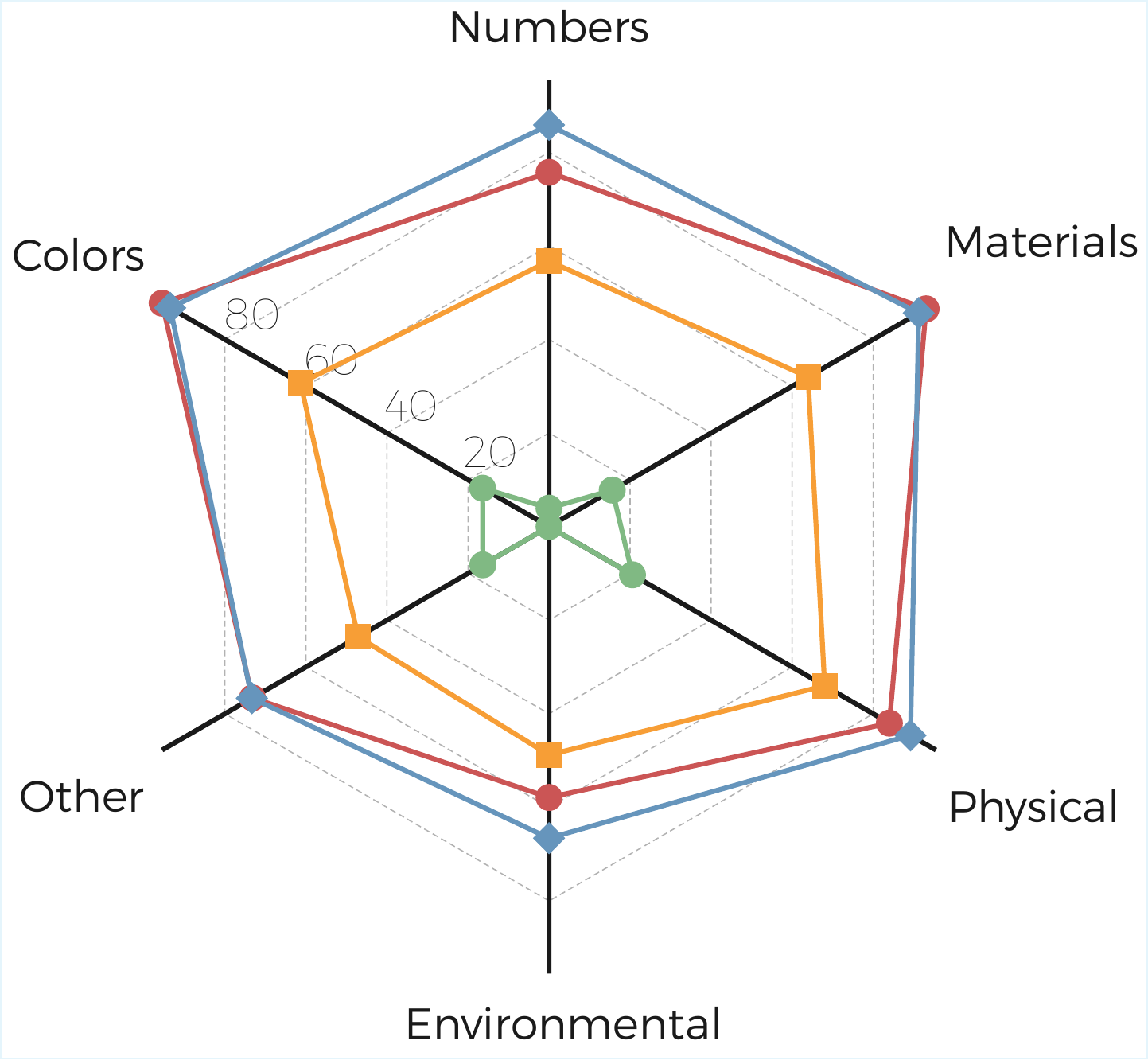}
    \caption{Category-specific performance on multiple-choice task. We show contradiction detection accuracy across object categories (\textbf{left}) and attribute categories (\textbf{right}) for four representative models.}
    \label{fig:multiple_choice_radar_plots}
\end{figure*}

\textit{(i) Performance hierarchy}.  There is a clear performance gap between closed- and open-source models.
Top-tier models (GPT 5 and Gemini 2.5 Pro) demonstrate strong conflict detection capabilities, correctly identifying conflicts 86.78\% and 88.48\% of the time, respectively. Their strong performance also validates that our dataset contains well-constructed, unambiguous samples with clear correct answers and confirms the task is solvable given sufficient reasoning capabilities. In contrast, most open-ended models struggle significantly, with many achieving less than 5\% accuracy, revealing a substantial capability gap in cross-modal reasoning. Notably, InstructBLIP-T5-xxl performs best among open-ended models with 63.87\% contradiction detection, though still far below the top-tier models. 

\textit{(ii) Modality bias patterns}.
When failing to detect contradictions, models exhibit distinct bias patterns. Closed-source models GPT-4.1 Mini (69.57\%) and Gemini 2.5 Flash Lite (74.98\%) show strong image bias. Similarly, LLaVa-OneVision-7b, Qwen3VL-30b and Qwen3VL-8b demonstrate even stronger image preference (75.42-81.45\%), while InternVL 1.5 and LLaVa-1.5 favor text-grounded responses (52.40\% and 44.68\%, respectively). The Qwen family shows bias evolution: Qwen2VL-7b is balanced, whereas Qwen2.5VL and Qwen3VL shift strongly to image preference, suggesting architecture, training, and scale influence modality dominance.

\textit{(iii) Scaling effects.} In the Qwen family, larger models do not consistently improve conflict detection: while Qwen2.5VL-32b (16.10\%) outperforms the 7b variant (7.79\%), Qwen3VL-30b (3.81\%) underperforms the 8b model (7.13\%). This suggests that scale alone does not guarantee better conflict detection.


\textit{(iv) Instruction following challenges.}
Several open-ended models exhibit high ``Incorrect'' response rates  (Phi3-vision-128k at 59.63\%, MiniGPT4-7b at 52.30\%), indicating difficulty in generating responses that match the required format. Notably, more recent models (Qwen2.5VL-32b, Qwen3VL, LLaVa-OneVision) have no incorrect responses, suggesting improved instruction-following capabilities in newer architectures. 

\textbf{Category-specific analysis.} 
To obtain a more fine-grained perspective of the performance differences, we analyze performance across object and attribute categories using four representative models. Fig.~\ref{fig:multiple_choice_radar_plots} presents the conflict detection accuracy (see Tables~\ref{tab:multiple_choice_attribute_categories} and \ref{tab:multiple_choice_object_categories} in App.~\ref{app:sec:experiments}). 

\textit{Object category patterns. }
Top--tier models perform best on Animals (GPT-5 with 97.32\%, Gemini 2.5 Pro with 98.65\%) and weakest on Household items ($\sim$78\%). InstructBLIP-T5-xxl maintains consistent 60-77\% performance across categories, while InternVL 1.5 performs poorly overall, with slight gains on Vehicles and Food.

\textit{Attribute category patterns. }
Environmental attributes prove most challenging for all models, while Colors are more easily detected. Numbers reveal an interesting capability difference between the leaders: Gemini 2.5 Pro substantially outperforming GPT-5 (85.83\% vs 75.70\%). InternVL 1.5 struggles across all attributes, while InstructBLIP-T5-xxl maintains moderate performance.

\subsection{Open-ended question answering}

 We evaluate model performance on the open-ended QA task, where models generate free-form responses to the same questions used in the multiple-choice evaluation. 

 \textbf{Evaluation protocol.}
 We use relaxed string matching (App.~\ref{app:subsec:relaxed_matching_open_ended}) to classify responses into  ``Conflict'', ``Incorrect'', image-grounded and text-grounded categories (Table~\ref{tab:open_ended_scores} in App.~\ref{app:sec:experiments}). LLM-as-a-judge evaluation with Gemini 2.5 Pro and GPT-5 confirms high agreement, validating the robustness of the evaluation (Table~\ref{tab:llm_judge_evaluation} in App.~\ref{app:sec:experiments}).

 \textbf{Performance evaluation.} Table~\ref{tab:open_ended_scores} presents conflict detection and incorrect response rates across models. Leading closed-source models maintain strong performance, with Gemini 2.5 Pro showing improved performance in the open-ended format (91.59\% vs 88.45\% in multiple-choice), while GPT-5 exhibits a slight decrease (81.16\% vs 86.78\%) primarily due to increased ``Incorrect'' (empty) responses. Notably, GPT-4.1 Mini shows improved performance (39.93\%) compared to its multiple-choice results (13.56\%). 
 In contrast, open-source models perform poorly, often achieving near-zero performance. Among exceptions, Qwen3VL models perform best (20.49\% and 19.80\%), followed by InternVL-1.5 (18.76\%) and Qwen2.5VL (17.13\%). These results indicate that current open-source models lack fundamental reasoning capabilities for cross-modal conflict detection. While instruction-following improves across generations: mPLUG-Owl-1 (61.61\% incorrect) vs.~mPLUG-Owl-2 (0.94\%), the format compliance advancement does not translate to improved conflict detection.

 \begin{table}[ht]
    \centering
    \caption{Percentage of detected conflicts and incorrect-format outputs on the open-ended task.
    }
    \resizebox{0.89\columnwidth}{!}{
        \begin{tabular}{lcc}
        \toprule
        \textbf{Model} & \textbf{Conflict} ($\uparrow$) & \textbf{Incorrect} ($\downarrow$) \\
        \midrule
        GPT 5 & $81.16_{\pm 1.10}$ & $8.24_{\pm 0.74}$ \\
        GPT 4.1 Mini & $39.93_{\pm 1.34}$ & $5.53_{\pm 0.65}$ \\
        Gemini 2.5 Pro & $91.59_{\pm 0.77}$ & $0.31_{\pm 0.16}$ \\
        Gemini 2.5 Flash Lite & $20.94_{\pm 1.14}$ & $9.32_{\pm 0.82}$ \\
        \midrule
        InternVL 1.5 & $18.76_{\pm 1.10}$ & $6.33_{\pm 0.67}$ \\
        mPLUG-Owl-1 & $6.56_{\pm 0.69}$ & $61.61_{\pm 1.35}$ \\
        mPLUG-Owl-2 & $1.41_{\pm 0.33}$ & $0.94_{\pm 0.27}$ \\
        LLaVa-1.5-7b & $0.00$ & $4.98_{\pm 0.59}$ \\
        LLaVa-OneVision-7b & $0.08_{\pm 0.08}$ & $15.73_{\pm 1.00}$ \\
        Qwen2VL-instruct-7b & $0.38_{\pm 0.17}$ & $10.01_{\pm 0.84}$ \\
        Qwen2.5VL-instruct-7b & $17.13_{\pm 1.07}$  & $7.37_{\pm 0.72}$   \\
        Qwen3VL-instruct-8b & $20.49_{\pm 1.12}$  &  $7.06_{\pm 0.73}$ \\
        Qwen3VL-instruct-30b & $19.80_{\pm 1.11}$ & $8.24_{\pm 0.81}$ \\
        \bottomrule
        \end{tabular}
    }
    \label{tab:open_ended_scores}
\end{table}
    \vspace{-1.2em}
\begin{figure}[h]
    \centering
    \includegraphics[width=0.92\columnwidth]{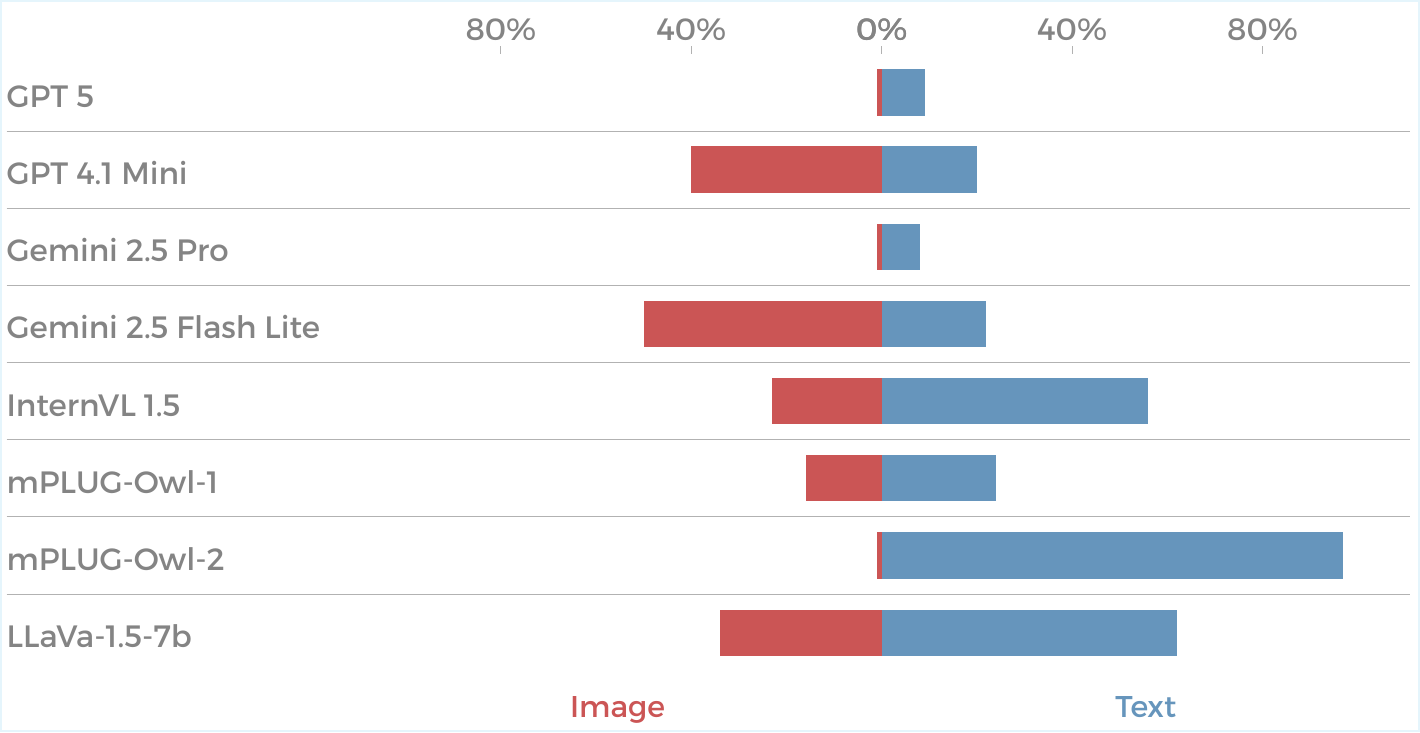}
    \caption{Modality preference on the open-ended task.}
    \label{fig:modality_preference_open_ended}
\end{figure}
\begin{table*}[ht]
\centering
\caption{
Finetuning results on the open-ended task.
We finetune LLaVa-1.5-7b and mPLUG-Owl-1 using $\sim$15k filtered and $\sim$30k raw samples. 
Columns report accuracy on conflicting and non-conflicting samples, overall accuracy across both, and the rate of incorrect outputs. Both models show substantial gains after finetuning. }
\resizebox{0.67\textwidth}{!}{
    \begin{tabular}{l cccc}
    \toprule
    \textbf{Model} & \textbf{Conflict} ($\uparrow$) & \textbf{No conflict} ($\uparrow$) & \textbf{Overall} ($\uparrow$) & \textbf{Incorrect} ($\downarrow$)\\
    \midrule
    LLaVa-1.5-7b & $0.00$ & $92.83_{\pm 0.73}$ & $46.44_{\pm 0.97}$ & $5.00_{\pm 0.61}$ \\
    LLaVa-1.5-7b-ft & $76.86_{\pm 1.16}$ & $91.21_{\pm 0.83}$ & $84.08_{\pm 0.74}$ & $0.47_{\pm 0.19}$ \\
    LLaVa-1.5-7b-ft-30k & $51.84_{\pm 1.40}$ & $88.71_{\pm 0.87}$ & $70.27_{\pm 0.90}$ & $8.21_{\pm 0.77}$ \\
    \midrule
    mPLUG-Owl-1 & $6.56_{\pm 0.70}$ & $31.62_{\pm 1.33}$ & $19.10_{\pm 0.78}$ & $61.63_{\pm 1.32}$ \\
    mPLUG-Owl-1-ft & $57.45_{\pm 1.41}$ & $63.13_{\pm 1.34}$ & $60.28_{\pm 0.97}$ & $7.09_{\pm 0.72}$ \\
    mPLUG-Owl-1-ft-30k & $50.61_{\pm 1.43}$ & $75.12_{\pm 1.22}$ & $62.85_{\pm 0.95}$ & $3.91_{\pm 0.53}$ \\
    \bottomrule
    \end{tabular}
}
\label{tab:open_ended_ft}
\end{table*}

Fig.~\ref{fig:modality_preference_open_ended} shows the distribution of image-- and text-grounded responses when models fail to detect contradictions. Several models demonstrate overwhelming text preference: mPLUG-Owl-2 shows extreme text bias (97.03\%), while InternVL 1.5 and LLaVa-1.5-7b also lean heavily textual cues (55.51\% and 62.10\% respectively). In contrast, GPT-4.1 Mini and Gemini 2.5 Flash Lite shows moderate image preference (40.39\% and 49.73\%). See Table~\ref{tab:open_ended_object_categories} and \ref{tab:open_ended_attribute_categories} in App.~\ref{app:sec:experiments} for category-specific analysis. 

\subsection{LoRA finetuning}
\label{subsec:lora_finetuning_main}

\citet{deng2025words} show that supervised finetuning effectively improves robustness to multimodal inconsistencies. Following this approach, we finetune the two worst-performing models, LLaVa-1.5-7b and mPLUG-Owl, on conflict detection examples. Implementation details are in App.~\ref{app:subsec:lora_finetuning}, with additional experiments in App.~\ref{app:sec:fine_tuning}.

\textbf{Training data. }
We construct $\sim$30k raw training samples following the pipeline described in \S\ref{subsec:dataset_construction}. After applying quality control measures (\S\ref{subsec:quality_controls}), which remove ambiguous samples, we retain $\sim$15k high-quality examples. To study data quality vs. quantity effects, we compare finetuning on the $\sim$30k full set vs. the $\sim$15k filtered subset. 

\textbf{Evaluation metrics.} We evaluate: i) conflict ($\uparrow$): contradiction detection rate on the \bench~test set; ii) no conflict ($\uparrow$): performance on samples with consistent image-text pairs (using original captions) to measure false positive rates; iii) overall ($\uparrow$): true positive rate across both contradiction and non-contradiction cases; and iv) incorrect ($\downarrow$): frequency of invalid outputs on \bench~that fail to match with either contradiction or modality-specific answers.

\textbf{Results and analysis.} Table~\ref{tab:open_ended_ft} shows that finetuning significantly boosts contradiction detection for both LLaVa-1.5-7b and mPLUG-Owl-1. 
For LLaVa, performance jumps from 0.00\% to \textbf{76.86\%} on contradiction samples with filtered training data, while maintaining strong accuracy on non-contradiction inputs (91.21\%). Error rates also drop sharply (5.00\% $\rightarrow$ 0.47\%). Training on unfiltered $\sim$30k data yields lower contradiction accuracy (51.84\%) and a higher error rate (8.21\%), highlighting that data quality outweighs quantity. mPLUG-Owl-1 also improves substantially, from 6.56\% to 57.45\% contradiction accuracy, with error rates reduced from 61.63\% to 7.09\%. Interestingly, unfiltered training gives higher non-contradiction accuracy (75.12\% vs 63.13\%). Overall, these results confirm that targeted finetuning enables open-ended models to acquire robust multimodal conflict detection capabilities.

\section{Conclusion}

We introduced \bench, the first large-scale diagnostic benchmark for spontaneous cross-modal conflict detection in dual ground truth scenarios. Unlike traditional datasets that treat a single modality -- typically the image -- as ground truth, \bench~requires models to critically evaluate both visual and textual information as equally valid sources, and detect inconsistencies. Through our systematic construction pipeline, we generated controlled contradictions, paired with targeted evaluation questions and carefully validated answer sets. Our comprehensive experiments reveal a stark performance gap between leading closed-source models and open-source models, expose systematic modality biases across model architectures, and demonstrate that targeted finetuning with high-quality data can substantially enhance conflict detection capabilities.

\textbf{Limitations and future work.}
While \bench~provides a controlled setting for studying multimodal inconsistencies, it has several limitations. First, it focuses on object-- and attribute--level contradictions, rather than spatial, relational, temporal or causal conflicts. This design choice is deliberate:  before evaluating whether models can detect complex inconsistencies, we must first establish whether they can perform basic cross-modal comparison on directly observable properties. Our results justify this approach: since most of the evaluated open-source models fail to detect ``simple'' mismatches, introducing more complex conflicts would conflate multiple failure modes. 
Second, building \bench~from everyday scenes in MS COCO ensures broad visual coverage and diversity across 80 categories and varied contexts, however extending it to specialized domains like medical, scientific, or technical content remains an important direction for future work. Finally, although some synthetic examples may contain lower quality outputs despite multi-stage filtering, synthetic generation offers several key benefits: precise isolation of single contradictory elements, systematic categorization and fine-grained experimental analysis. Future work could further strengthen reliability through enhanced validation, such as cross-model agreement checks or multi-annotator verification.

Despite these limitations, \bench~offers a unique diagnostic value:  controlled evaluation of fundamental capabilities, systematic modality bias quantification, and fine-grained category analysis. We hope \bench~serves as a foundational tool for developing models that critically reason about multimodal input consistency, paving the way for more trustworthy and reliable multimodal AI systems.

{
    \small
    \bibliographystyle{ieeenat_fullname}
    \bibliography{main}
}

\clearpage
\appendix

\twocolumn[
\begin{center}
    \Large\textbf{Supplementary Material \\ \bench: A Benchmark for Cross-Modal Contradiction Detection}
\end{center}
\vspace{1em}
]

The supplementary material is organized as follows:
\begin{itemize}
    \item Implementation details (App.~\ref{app:sec:implementation_details})
    \item The prompts for data generation, quality check and model evaluation (App.~\ref{app:sec:prompts})
    \item Additional experiments and results on \bench (App.~\ref{app:sec:experiments})
    \item Comprehensive list of the changed words (from original to conflict in caption) categorized into objects/attributes (App.~\ref{app:sec:object_categories} and App.~\ref{app:sec:attribute_categories})
    \item Qualitative examples of \bench~question types (App.~\ref{app:sec:qualitative_examples})
    \item Details about the human validation of the test set (App.~\ref{app:sec:human_validation})
    \item Broader impact (App.~\ref{app:sec:broader_impact})
\end{itemize}

\section{Implementation details}
\label{app:sec:implementation_details}

\subsection{Relaxed string matching for multiple choice tasks}
\label{app:subsec:relaxed_matching_mutliple_choice}

Our evaluation protocol employs a hierarchical matching approach that accommodates various response formats. The matching process follows three sequential steps:
\begin{enumerate}[label=(\roman*), leftmargin=*]
    \item \textbf{Bracketed choices (strict matching)}: Responses like (A), (B), (C), (D) are matched directly to answer categories. If multiple bracketed options appear, the response is marked incorrect.  
    \item \textbf{Single letters}: Responses like A--D (without brackets) are normalized and matched to the expected choices.  
    \item \textbf{Keyword matching}: If neither of the above applies, we check whether the response contains the exact answer text as a standalone word (case-insensitive, using word boundaries).  
\end{enumerate}

\subsection{Relaxed string matching for open-ended tasks}
\label{app:subsec:relaxed_matching_open_ended}

The open-ended question answering task presents models with conflicting image-text pairs and asks them to provide free-form responses explaining what they observe. We classify model responses into four mutually exclusive categories.
\begin{prompt}{Four response categories}
\begin{enumerate}
    \item Conflict detection: responses that correctly identify the contradiction between image and text content.
    \item Image-grounded: responses that describe or rely primarily on visual information. 
    \item Text-grounded: responses that align with or prioritize textual information. 
    \item Incorrect: responses that are unrelated to the content, incoherent, or fail to address the question meaningfully.
\end{enumerate}
\end{prompt}

We implement a multi-step normalization pipeline to handle the natural variability in open-ended responses:
\begin{itemize}
    \item Tokenization: extract word tokens and convert to lowercase.
    \item Articles removal: filter out articles (the, a, an).
    \item Number normalization: map numeric tokens to their word equivalents (e.g., ``1'' → ``one'').
    \item Word stripping: remove common linguistic variations using a predefined mapping (e.g., ``wooden'' → ``wood'', ``brightly'' → ``bright'').
    \item Stemming: apply Porter stemming to reduce words to their root forms, handling morphological variations.
\end{itemize}

For each response category, we perform substring matching on the normalized text:
\begin{itemize}
    \item Contradiction detection: check for presence of stemmed versions of ``conflict'' or ``contradict'' using word boundary matching.
    \item Modality-specific responses: match against normalized versions of the expected image-only or text-only answers for each sample.
    \item Boundary matching: use regex word boundaries to ensure whole-word matches and avoid partial matches.
\end{itemize}

The automated evaluation system was manually validated on a subset of responses to ensure classification accuracy.

\subsection{Finetuning with LoRA}
\label{app:subsec:lora_finetuning}

To avoid the strong bias of flagging conflicts, we pick "conflicting caption" or "original caption" with equal probability during finetuning.
The LoRA finetuning is conducted for LLaVa-1.5-7b and mPLUG-Owl-1, both in 1 epoch. Following their official repository \footnote{LLaVa \url{https://github.com/haotian-liu/LLaVA}, mPLUG-Owl-1 \url{https://github.com/X-PLUG/mPLUG-Owl}}, we adopt bf16 and gradient checkpointing for efficiency. All experiments are conducted in a single A100-64GB. Check LoRA settings and training hyperparameters in Table~\ref{tab:lora}.

\begin{table}[ht]
\centering
\caption{LoRA hyperparameters for multimodal conflict detection finetuning.}
\begin{tabular}{lcc}
    \toprule
     & \textbf{LLaVa-1.5-7b}  & \textbf{mPLUG-Owl-1}  \\
    \midrule
    LoRA-$r$  & 128 & 8 \\
    LoRA-$\alpha$ & 256 & 32 \\
    dropout & 0.05 & 0.05 \\
    \midrule
    sequ. length & 2048 & 2048\\
    batch size & 16 & 4 \\
    learning rate & 2e-5 & 2e-5 \\
    scheduler & cosine & cosine \\
    warmup & 0.03 ratio & 50 steps \\
\bottomrule
\end{tabular}
\label{tab:lora}
\end{table}

\section{Prompts}
\label{app:sec:prompts}

\subsection{Data generation}
\label{app:sec:prompts_data_generation}

In this section, we present the prompts employed for data generation. These prompts were carefully designed to elicit high-quality outputs from the model while controlling for specific linguistic and visual attributes. 

\begin{prompt}{Generate conflicting caption and question}
You are an expert image caption editor and question generator. Your task is to modify existing image captions and then create subtle questions based on your modifications. Given an original image caption, you need to perform the following four steps:

\textbf{Step 1: Create a conflicting caption.}
\begin{itemize}
    \item Take the provided original caption.
    \item Identify \emph{one} key element (either a specific object or an attribute of an object, like its color, number, shape, material, or texture).
    \item Change \emph{only this one element} to create a subtle, but noticeable, conflict or discrepancy. The rest of the caption must remain identical to the original.
    \item Ensure the conflict is a plausible, though incorrect, alternative (e.g., ``red car'' to ``blue car,'' not ``red car'' to ``flying car'').
    \item Do not change words that have binary states (e.g., man--woman, open--close, dark-light, indoor--outdoor).
\end{itemize}

When changing an attribute:
\begin{itemize}
    \item Only change \textbf{objective attributes} such as \textbf{color, number, shape, material, or texture}.
    \item Do not change \textbf{subjective or ambiguous attributes} such as beautiful, small, large, big, medium, moderate, modern, young, old, fast, slow, elegant, scary, tall, short, etc.
\end{itemize}

\textbf{Step 2: Track the changed words.}
\begin{itemize}
    \item Identify the exact word(s) that were changed from the original caption.
    \item Record both the original word(s) and the replacement word(s).
\end{itemize}

\textbf{Step 3: Identify the type of change.}
\begin{itemize}
    \item Determine whether the change made in Step 1 was to an ``object'' (e.g., ``cat'' changed to ``dog'') or an ``attribute'' (e.g., ``white'' cat changed to ``black'' cat).
\end{itemize}

\textbf{Step 4: Generate a subtle question.}
\begin{itemize}
    \item Based on your \emph{newly created conflicting caption}, formulate a question.
    \item This question must subtly hint at the conflicting element without directly stating that something is wrong or different.
    \item The question should encourage the user to focus on the changed element.
\end{itemize}

Here is an example. Provide your responses in the exact JSON format shown:

\begin{lstlisting}
User: "A fluffy white cat sitting on a red couch."
Model:
{
  "conflicting_caption": "A fluffy black cat sitting on a red couch.",
  "question": "What color is the cat sitting on a couch?",
  "change_type": "attribute",
  "changed_words": {
    "original": "white",
    "conflicting": "black"
  }
}
\end{lstlisting}
\end{prompt}

\begin{prompt}{Generate multiple choice answers}
Your task is to write \textbf{three answer choices} for a multiple-choice question that highlights a subtle conflict between two captions.

\subsection*{Instructions}

Generate the following three answer options:

\begin{enumerate}
    \item \textbf{image\_only\_answer}: The answer that fits the original caption.
    \item \textbf{text\_only\_answer}: The answer that matches the conflicting caption.
    \item \textbf{irrelevant\_but\_plausible}: A plausible distractor that doesn't appear in either the original or the conflicting caption, but is contextually reasonable. 
    \begin{itemize}
        \item Make sure the distractor is NOT ambiguous or vague (avoid words such as \emph{several, afternoon, medium, moderate, thing, stuff}).
    \end{itemize}
\end{enumerate}

Ensure that all answers are:
\begin{itemize}
    \item Concise (preferably 1--3 words),
    \item Mutually exclusive,
    \item Plausible in context, but not synonymous to each other.
\end{itemize}

Only output a JSON with the following fields:
\begin{lstlisting}
{
  "image_only": "...",
  "text_only": "...",
  "irrelevant_but_plausible": "..."
}
\end{lstlisting}
\end{prompt}

\subsection{Data filtering}
\label{app:sec:prompts_data_filtering}

This section describes the data filtering procedures applied to ensure the quality and consistency of the generated dataset. We perform multiple checks, including attribute verification, object validation, answer consistency, and question clarity. Each filtering step is designed to identify and remove entries that are ambiguous, irrelevant, or inconsistent, thereby maintaining the reliability of the dataset for downstream evaluation and analysis. 

\begin{prompt}{Attribute check}
You are given two words: \texttt{original} and \texttt{conflicting}. Perform the following two steps:

\begin{enumerate}
    \item Decide if each word is an \textbf{attribute} (a descriptive property, e.g., red, tall, beautiful) or \textbf{not an attribute} (an object, e.g., car, tree).
    \item If both are attributes, classify them as:
    \begin{itemize}
        \item \textbf{Objective} = measurable, factual, observable (e.g., red, square, wooden, three).
        \item \textbf{Subjective} = opinion-based or interpretive (e.g., beautiful, small, moderate, large, big, medium, modern, young, old, fast, slow, elegant, scary, tall, short, stylish, fancy, cheap, impressive).
    \end{itemize}
\end{enumerate}

Only output a JSON with the following fields:
\begin{lstlisting}
{
  "change_is_attribute": "Yes/No",
  "change_is_objective": "Yes/No"
}
\end{lstlisting}
\end{prompt}

\begin{prompt}{Object check}
You are given two words: \texttt{original} and \texttt{conflicting}.  

Your task is to check the quality of the conflicting word in relation to the original:

\begin{enumerate}
    \item \textbf{Object check:} Determine whether each word is an object (a tangible or identifiable thing/entity, e.g., car, apple, chair).  
    \begin{itemize}
        \item Are the two words objects?
    \end{itemize}
    
    \item \textbf{Synonymy check:} Is the conflicting object a synonym or near-synonym of the original?
    
    \item \textbf{Ambiguity check:} Is the conflicting object ambiguous or vague (e.g., ``thing'', ``object'', ``stuff'')?
    
    \item \textbf{Contextual relevance:} Does the conflicting object make sense in the same scene as the original?
\end{enumerate}

Only output a JSON with the following fields:

\begin{lstlisting}
{
  "change_is_object": "Yes/No",
  "change_is_synonym": "Yes/No",
  "change_is_ambiguous": "Yes/No",
  "change_is_relevant": "Yes/No"
}

\end{lstlisting}
\end{prompt}

\begin{prompt}{Answers check}
You are given three words/phrases. For each word/phrase, check the following:

\begin{enumerate}
    \item \textbf{Synonymy check:} Is one of the words/phrases a synonym or near-synonym of the other two?
    
    \item \textbf{Ambiguity check:} Is any of the words/phrases ambiguous or vague (e.g., ``several'', ``afternoon'', ``medium'', ``thing'')?
    
    \item \textbf{Contextual relevance:} Are all words/phrases contextually relevant and objective (not subjective or off-topic)?
    
    \item \textbf{Visual check:} Can each word/phrase be directly observed in an image?  
    Examples of visual words include:
    \begin{itemize}
        \item Attributes of objects (number, color, shape, size, material)  
        \item Object categories (car, chair, dog)  
        \item Spatial relations (on top of, next to)  
        \item Scenes (beach, kitchen)
    \end{itemize}
    Examples of non-visual words include:
    \begin{itemize}
        \item Temporal concepts (afternoon, tomorrow)  
        \item Abstract states (freedom, happiness)  
        \item Non-observable attributes (brand, taste, temperature)  
        \item Subjective labels (beautiful, boring, large)
    \end{itemize}
\end{enumerate}

\begin{lstlisting}
{
  "ans_is_synonym": "Yes/No",
  "ans_is_ambiguous": "Yes/No",
  "ans_is_relevant": "Yes/No",
  "ans_is_visible": "Yes/No"
}
\end{lstlisting}

\end{prompt}

\begin{prompt}{Questions check}
You are given a question, a set of changed words, and three possible answers. Your task is to check the question based on three criteria:

\begin{enumerate}
    \item \textbf{Ambiguity check:} Is the \textbf{question} clear, specific, and unambiguous?
    
    \item \textbf{Focus check:} Does the question explicitly ask about the \textbf{changed words}?  
    \textit{Examples:}  
    \begin{itemize}
        \item Question: ``What is the color of something?'' Changed words: green, blue $\rightarrow$ Output: Yes  
        \item Question: ``How many items are there?'' Changed words: three, four $\rightarrow$ Output: Yes
    \end{itemize}
    
    \item \textbf{Answerability:} Are all three candidate \textbf{answers} semantically and contextually compatible with the question?  
    \textit{Examples:}  
    \begin{itemize}
        \item Question: ``What is the color?'' Answers: green, blue, red $\rightarrow$ Output: Yes  
        \item Question: ``What is the gender?'' Answers: man, woman, child $\rightarrow$ Output: No
    \end{itemize}
\end{enumerate}

Only output a JSON object with the following fields:

\begin{lstlisting}
{
  "question_is_clear": "Yes/No",
  "question_is_focused": "Yes/No",
  "question_is_answerable": "Yes/No"
}
\end{lstlisting}

\end{prompt}

\subsection{Model evaluation}
\label{app:sec:prompts_model_evaluation}

This section outlines the procedures used to evaluate model performance across different tasks. We describe the evaluation instructions for multiple-choice and open-ended questions, as well as the framework in which a large language model (LLM) acts as a judge to assess response quality. 

\begin{prompt}{Multiple choice instruction}
You are given an image and a textual description of the image.  

Use \emph{both} the visual and textual information to answer the multiple-choice question below as accurately as possible.  

\begin{itemize}
    \item Choose the option that is best supported by \textbf{both sources}.
    \item If the image and the description contradict each other, or if one is missing key information, select the option ``Conflicting information – cannot answer.''
    \item Answer only with the letter corresponding to your choice in parentheses: (A), (B), (C), or (D).
    \item Do not include any other text.
\end{itemize}
\end{prompt}

\begin{prompt}{Open-ended instruction}

You are given an image and a textual description of the image.  

Use \emph{both} the visual and textual information to answer the question below as accurately as possible.  

\begin{itemize}
    \item Only answer if your response can be clearly supported by \textbf{both sources}.
    \item If the image and the description contradict each other, or if one is missing key information, output ``Conflicting information – cannot answer''.
    \item Otherwise, provide the answer (less than 15 words). 
\end{itemize}
\end{prompt}

\begin{prompt}{LLM-as-a-judge}
 You are an evaluator. Given two reference answers (\texttt{image\_only} and \texttt{text\_only}) and a model prediction, decide which category the prediction belongs to:

\begin{enumerate}
    \item \textbf{IMAGE} — if the prediction semantically matches the \texttt{image\_only} answer.
    \item \textbf{TEXT} — if the prediction semantically matches the \texttt{text\_only} answer.
    \item \textbf{CONFLICT} — if the prediction explicitly refers to a contradiction, conflict, or states that both cannot be true.
    \item \textbf{NONE} — if the prediction matches neither answer and does not indicate a conflict.
\end{enumerate}

Ignore minor differences in phrasing, synonyms, plural/singular forms, or capitalization. Return only one label: \textbf{IMAGE}, \textbf{TEXT}, \textbf{CONFLICT}, or \textbf{NONE}.

\textbf{Examples:}
\begin{verbatim}
Image-only answer: polar bear
Text-only answer: brown bear
Prediction: Brown bear
Output: TEXT

Image-only answer: Black
Text-only answer: Blue
Prediction: Conflicting information.
Output: CONFLICT

Image-only answer: dog
Text-only answer: cat
Prediction: dog
Output: IMAGE

Image-only answer: red
Text-only answer: green
Prediction: yellow
Output: NONE
\end{verbatim}
\end{prompt}

\section{Experiments}
\label{app:sec:experiments}
In this section we report additional experiments and detailed analysis of model performance on \bench. We provide comprehensive results across both multiple-choice and open-ended evaluation formats, including category-specific breakdowns and validation of our evaluation methodology through LLM-as-a-judge assessment.

\subsection{Multiple choice question answering}
\label{app:subsec:multiple_choice}

\paragraph{Performance evaluation. } Table~\ref{tab:multiple_choice_results_relaxed_matching} shows percentage of predictions matching the respective answer for the multiple-choice question answering task using relaxed string matching for evaluation. This format tests models' ability to recognize conflicts when provided with explicit options, including the correct "Conflicting information – cannot answer" choice.

\begin{figure}[ht!]
    \centering
    \includegraphics[width=1\linewidth]{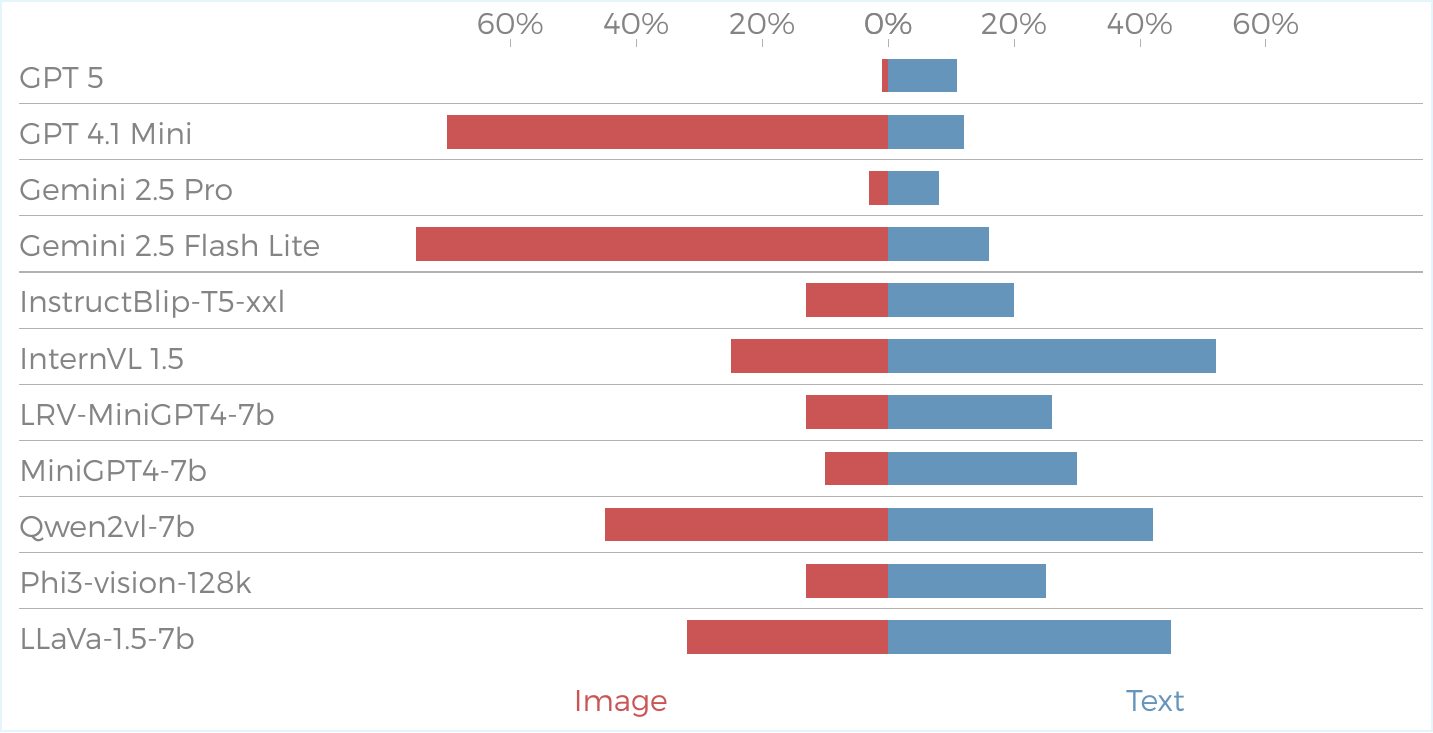}
    \caption{Modality preference patterns in multiple-choice QA. Most models exhibit systematic biases, favoring either visual (red) or textual (blue) information when faced with contradictions. }
    \label{fig:modality_preference_multiple_choice}
\end{figure}

Figure~\ref{fig:modality_preference_multiple_choice} depicts the modality preference of various models, revealing systematic biases toward either visual or textual information. Leading closed-source models (GPT-5, Gemini 2.5 Pro) show minimal bias, while their lighter variants (GPT-4.1 Mini, Gemini Flash Lite) demonstrate strong image preference. Open-source models show varying degrees of modality preference, with some strongly favoring text (InterVL-1.5, LLaVa-1.5-7b) and others showing more balanced distributions.

\begin{table*}[ht!]
\centering
\caption{Percentage of predictions matching the respective answer for the multiple-choice question answering task using relaxed string matching. Last column denotes cases where no match with any of the answers was found.}
\resizebox{0.7\textwidth}{!}{
\begin{tabular}{l cccccc}
    \toprule
    \textbf{Model} & \textbf{Conflict} ($\uparrow$) & \textbf{Image} ($\downarrow$)& \textbf{Text} ($\downarrow$)& \textbf{Distractor} ($\downarrow$)& \textbf{Incorrect} ($\downarrow$)\\
    \midrule
    Phi3-vision-128k & $1.29_{\pm 0.28}$ & $27.51_{\pm 1.10}$ & $53.65_{\pm 1.35}$ & $1.19_{\pm 0.28}$ & $16.30_{\pm 0.96}$ \\
    MiniGPT4-7b & $2.18_{\pm 0.38}$ & $23.50_{\pm 1.13}$ & $51.11_{\pm 1.27}$ & $9.88_{\pm 0.75}$ & $20.50_{\pm 1.04}$ \\
    mPLUG-Owl-2 & $0.00$ & $0.72_{\pm 0.22}$ & $77.92_{\pm 1.06}$ & $0.20_{\pm 0.12}$ & $21.10_{\pm 1.05}$ \\
    LRV-MiniGPT4-7b & $2.16_{\pm 0.39}$ & $24.10_{\pm 1.06}$ & $42.77_{\pm 1.24}$ & $10.69_{\pm 0.78}$ & $30.41_{\pm 1.21}$ \\
    InternVL 1.5 & $17.05_{\pm 0.99}$ & $26.35_{\pm 1.12}$ & $53.65_{\pm 1.31}$ & $2.34_{\pm 0.38}$ & $0.80_{\pm 0.23}$ \\
    InstructBlip-T5xxl & $63.86_{\pm 1.23}$ & $12.67_{\pm 0.84}$ & $19.89_{\pm 1.02}$ & $3.30_{\pm 0.47}$ & $0.06_{\pm 0.06}$ \\
    LLaVa-1.5-7b & $0.13_{\pm 0.09}$ & $38.98_{\pm 1.23}$ & $57.89_{\pm 1.29}$ & $3.09_{\pm 0.43}$ & $0.00$ \\
    Qwen2vl-instruct-7b & $1.27_{\pm 0.29}$ & $50.12_{\pm 1.31}$ & $47.64_{\pm 1.27}$ & $1.01_{\pm 0.25}$ & $0.00$ \\
    \bottomrule
    \end{tabular}
}
\label{tab:multiple_choice_results_relaxed_matching}
\end{table*}

\paragraph{Evaluation with non-conflicting samples.} 
To better understand the robustness of MM-LLMs in detecting multimodal contradictions, we evaluate models on data that includes both conflicting and non-conflicting samples. 
This experiment uses the same models and evaluation protocols but incorporates samples where the original caption (without modifications) is presented alongside the image, creating scenarios where no contradiction exists.  In these non-conflicting cases, models should respond with the image--grounded answer.

\begin{table*}[ht!]
\centering
\caption{Performance of various models on multiple-choice QA including conflicting and non-conflicting samples. \textit{Conflict} shows accuracy on contradictory samples, \textit{No conflict} on consistent samples, and \textit{Overall} is the true positive rate across both. }
\resizebox{0.5\textwidth}{!}{
    \begin{tabular}{l ccc}
    \toprule
    \textbf{Model} & \textbf{Conflict} ($\uparrow$) & \textbf{No conflict} ($\uparrow$) & \textbf{
    Overall} ($\uparrow$) \\
    \midrule
    \rowcolor{midgray}
    \multicolumn{4}{l}{\textit{\textbf{Strict string matching}}} \\
    InstructBlip-T5xxl & $63.54_{\pm 1.36}$ & $42.66_{\pm 1.34}$ & $53.18_{\pm 0.98}$ \\
    InternVL1.5 & $16.25_{\pm 1.02}$ & $92.51_{\pm 0.73}$ & $54.27_{\pm 0.95}$ \\
    MiniGPT4-7b & $2.32_{\pm 0.42}$ & $35.39_{\pm 1.33}$ & $18.80_{\pm 0.78}$ \\
    MiniGPT4-7b-LRV & $2.18_{\pm 0.42}$ & $36.01_{\pm 1.32}$ & $19.04_{\pm 0.77}$ \\
    Phi3-vision-128k & $0.94_{\pm 0.27}$ & $48.77_{\pm 1.40}$ & $24.80_{\pm 0.86}$ \\
    Qwen2vl-7b & $1.25_{\pm 0.30}$ & $94.55_{\pm 0.65}$ & $47.82_{\pm 0.98}$ \\
    LLaVa1.5-7b & $0.07_{\pm 0.08}$ & $63.84_{\pm 1.36}$ & $31.95_{\pm 0.92}$ \\
    \midrule
    \rowcolor{midgray}
    \multicolumn{4}{l}{\textit{\textbf{Relaxed string matching}}} \\
    InstructBlip-T5xxl & $63.59_{\pm 1.35}$ & $42.70_{\pm 1.42}$ & $53.17_{\pm 0.98}$ \\
    InternVL1.5 & $16.71_{\pm 0.99}$ & $92.98_{\pm 0.70}$ & $54.71_{\pm 1.01}$ \\
    MiniGPT4-7b & $2.34_{\pm 0.42}$ & $63.13_{\pm 1.33}$ & $32.62_{\pm 0.96}$ \\
    MiniGPT4-7b-LRV & $2.17_{\pm 0.41}$ & $55.71_{\pm 1.36}$ & $28.89_{\pm 0.87}$ \\
    Phi3-vision-128k & $1.08_{\pm 0.30}$ & $89.45_{\pm 0.83}$ & $45.13_{\pm 1.02}$ \\
    Qwen2vl-instruct-7b   & $1.31_{\pm 0.31}$ & $97.02_{\pm 0.47}$ & $49.04_{\pm 0.96}$ \\    
    LLaVa1.5-7b & $0.08_{\pm 0.08}$ & $90.44_{\pm 0.81}$ & $45.24_{\pm 0.99}$ \\
    \bottomrule
    \end{tabular}
}
\label{tab:multiple_choice_no_contr}
\end{table*}

We report the results of this experiment in Table~\ref{tab:multiple_choice_no_contr}, both with strict and relaxed string matching. The ``Conflict'' column shows accuracy on samples with visual-textual contradictions, ``No conflict'' column shows accuracy on samples with consistent information, and ``Overall'' represents the true positive rate across both conflicting and non-conflicting samples.
InternVL1.5, despite low conflict detection (16\%), achieves exceptional performance on non-conflicting samples (93\%), suggesting the model model can appropriately respond to non-conflicting information but fails at conflict identification. Conversely, InstructBlip-T5xxl shows stronger conflict detection (64\%) but weaker non-conflict performance (43\%).
LLaVa-1.5-7b achieves near-zero conflict detection but achieves moderate performance on non-conflicting samples (63.84\%).

Comparing the strict vs. relaxed string matching evaluation reveals that most models demonstrate minimal performance changes. However, several models show substantial improvements in non-conflict performance under relaxed matching. Models like LLaVa1.5-7b and Phi3-vision-128k appear to understand task requirements but struggle with strict answer formatting, leading to substantial underestimation of their capabilities under strict evaluation.

\paragraph{Category-specific performance analysis.} To understand how different types of contradictions affect performance, we analyze results across semantic categories.
Tables~\ref{tab:multiple_choice_attribute_categories} and \ref{tab:multiple_choice_object_categories} show an overview of the performance split across the object and attribute categories, respectively.
This breakdown reveals whether models struggle more with certain types of contradictions (e.g., color vs. environmental characteristics) and helps identify systematic weaknesses in multimodal reasoning capabilities.

\begin{figure*}[th!]
    \centering
    \begin{tabular}{ccc}
        \includegraphics[width=0.3\linewidth]{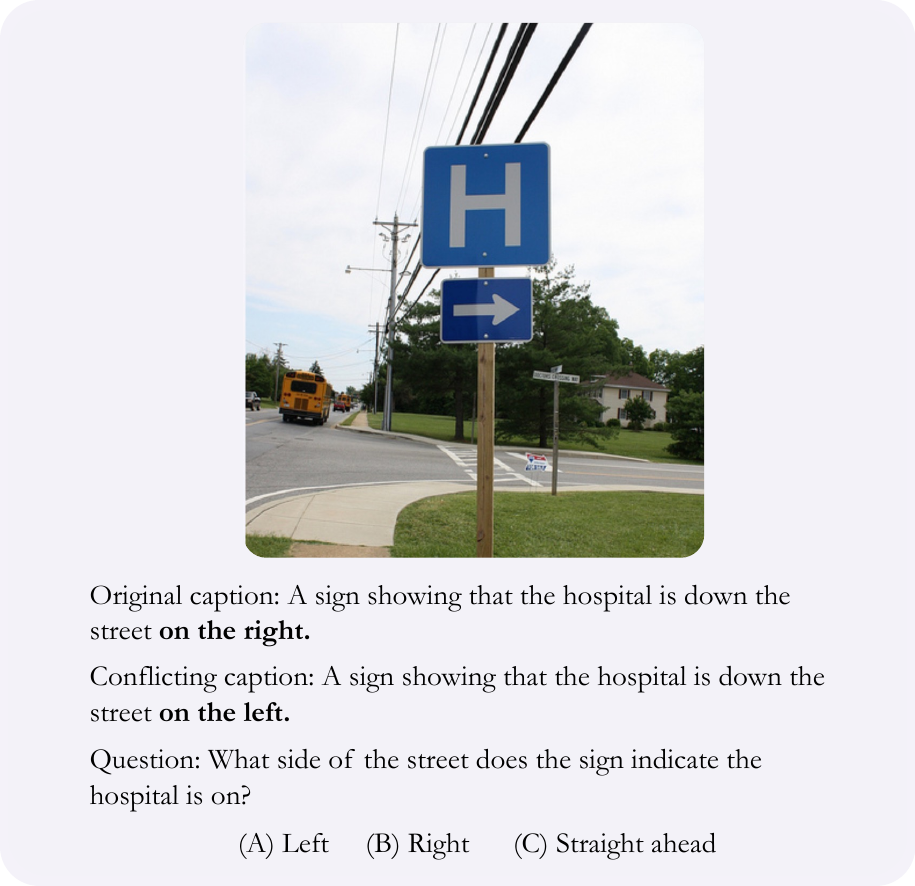} & 
        \includegraphics[width=0.3\linewidth]{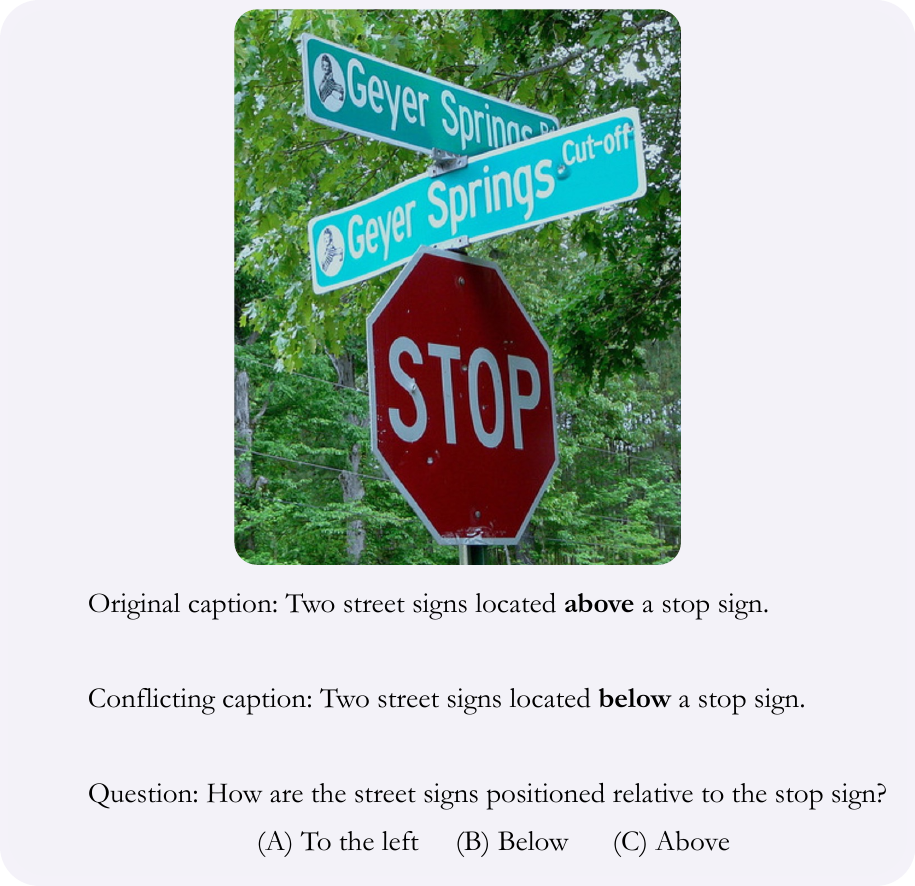} & 
        \includegraphics[width=0.3\linewidth]{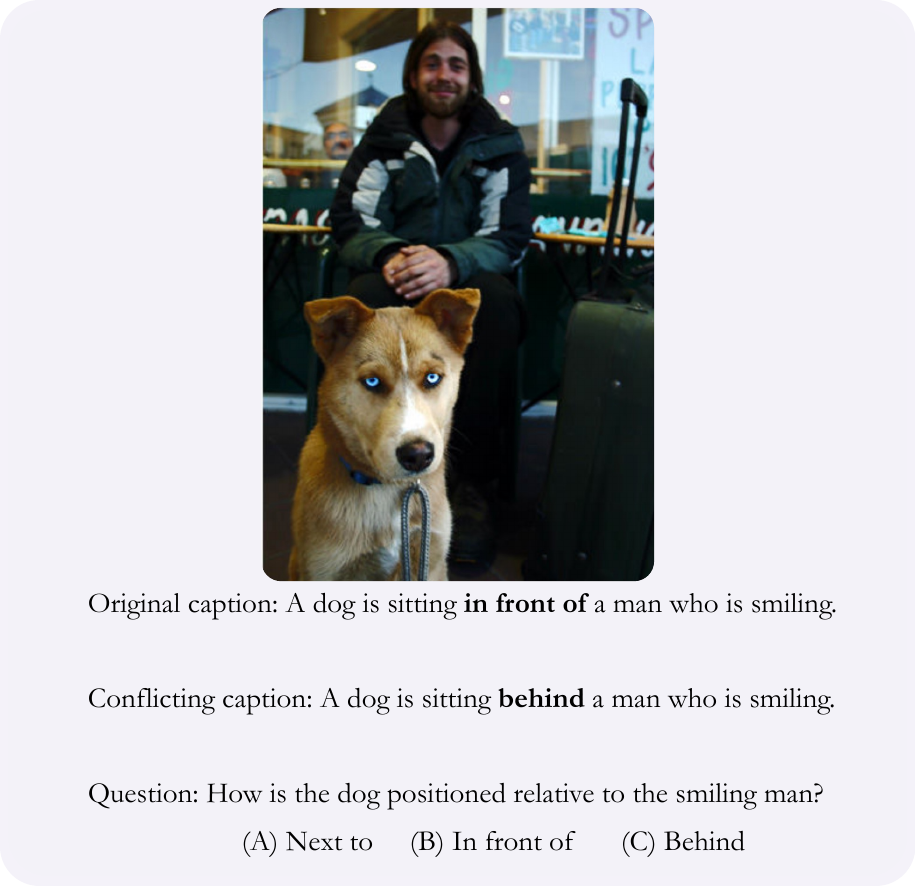} \\
        Left vs. right & Above vs. below & In front of vs. behind \\
    \end{tabular}
    \caption{Qualitative examples from the spatial conflict subset. We show left-right, above-below, and in-front-of/behind conflicts where captions contradict the spatial relationships visible in the images.}
    \label{fig:qualitative_examples_spatial}
\end{figure*}

\paragraph{Spatial conflicts.} While our main benchmark focuses on object- and attribute-level contradictions, we construct a small exploratory subset to assess whether models can detect spatial inconsistencies. We create 271 human-verified spatial conflict examples spanning three relation types: left-right positioning (151 samples), above-below vertical relationships (59 samples), and in-front-of/behind depth ordering (61 samples). These conflicts modify captions to reverse or contradict the spatial relationships present in images. The question and the distractor answer are generated with Gemini 2.5 Flash. Fig.~\ref{fig:qualitative_examples_spatial} shows representative examples from each spatial conflict type.

Table~\ref{tab:multiple_choice_results_spatial} shows that spatial conflicts are substantially harder than object- and attribute-level inconsistencies for all models. GPT-5 achieves 68.24\% spatial conflict detection, a notable drop from its 86.78\% performance on object and attribute conflicts (Table~\ref{tab:multiple_choice_results}). Open-source models show even bigger degradation: InstructBlip-T5xxl drops from 63.87\% on objects and attributes to 31.23\% spatial conflict detection, while InternVL1.5 nearly completely fails at spatial conflicts, achieving only 4.42\% detection compared to 16.71\% on object and attribute-level conflicts.

These results confirm our design rationale: if models cannot reliably detect ``simple'' object- and attribute-level mismatches, introducing spatial complexity only amplifies existing failure modes. Moreover, adding spatial conflicts conflates multiple failure modes—models could fail due to object recognition errors, inability to parse spatial relations from text, failures in visual grounding of entities, or weakness in comparing relational structures across modalities--undermining the diagnostic value of the benchmark.

\begin{table*}[ht]
\centering
\footnotesize
\caption{Percentage of predictions matching the respective answer on the multiple-choice task using \textbf{spatial} conflict types. The last column denotes cases where no match with any of the answers was found. The error bars$_{\pm}$ show standard deviation. 
}
\resizebox{0.7\textwidth}{!}{
    \begin{tabular}{l cccccc}
    \toprule
    \textbf{Model} & \textbf{Conflict} ($\uparrow$)& \textbf{Image} & \textbf{Text} & \textbf{Distractor} & \textbf{Incorrect} \\
    \midrule
    GPT 5 & $68.24_{\pm 2.92}$ & $1.46_{\pm 0.71}$ & $17.21_{\pm 2.32}$ & $0.00$ & $12.88_{\pm 2.08}$ \\
    \midrule
    InstructBlip-T5xxl & $31.23_{\pm 2.83}$ & $7.66_{\pm 1.64}$ & $43.62_{\pm 2.95}$ & $17.39_{\pm 2.34}$ & $0.00$ \\
    InternVL1.5 & $4.42_{\pm 1.26}$ & $21.84_{\pm 2.51}$ & $67.80_{\pm 2.89}$ & $4.09_{\pm 1.18}$ & $1.85_{\pm 0.85}$ \\

    \bottomrule
    \end{tabular}
}
\label{tab:multiple_choice_results_spatial}
\end{table*}

\begin{table*}[ht]
\centering
\caption{Object category performance breakdown for multiple choice QA. We report mean $\pm$ standard deviation. Results across five major categories reveal category-specific strengths and weaknesses in multimodal conflict detection.  Performance variations suggest that different object types pose varying difficulty levels for conflict detection, potentially due to visual saliency, semantic complexity, or training data distribution.}
\resizebox{0.8\textwidth}{!}{
    \begin{tabular}{lcccccc}
    \toprule
    \textbf{Model} & \textbf{Animals} & \textbf{Vehicles} & \textbf{Food} & \textbf{Sports} & \textbf{Household} & \textbf{Other} \\
    \midrule
    GPT 5 & $97.32_{\pm 1.33}$ & $93.91_{\pm 2.97}$ & $83.88_{\pm 3.92}$ & $89.07_{\pm 3.15}$ & $78.38_{\pm 3.25}$ & $77.58_{\pm 4.15}$ \\
    GPT 4.1 Mini & $3.47_{\pm 1.49}$ & $10.38_{\pm 3.75}$ & $14.89_{\pm 3.72}$ & $1.06_{\pm 1.07}$ & $8.54_{\pm 2.39}$ & $8.80_{\pm 2.75}$ \\
    Gemini 2.5 Pro & $98.65_{\pm 0.90}$ & $88.06_{\pm 4.03}$ & $88.56_{\pm 3.46}$ & $89.12_{\pm 3.22}$ & $77.74_{\pm 3.60}$ & $82.62_{\pm 3.66}$ \\
    Gemini 2.5 Flash Lite & $8.25_{\pm 2.28}$ & $6.12_{\pm 2.90}$ & $10.36_{\pm 3.38}$ & $3.37_{\pm 1.91}$ & $10.58_{\pm 2.60}$ & $4.84_{\pm 2.10}$ \\
    \midrule
    InstructBlip-T5xxl & $76.66_{\pm 3.36}$ & $71.70_{\pm 5.52}$ & $66.60_{\pm 4.98}$ & $60.82_{\pm 5.11}$ & $64.52_{\pm 3.85}$ & $59.88_{\pm 4.83}$ \\
    InternVL1.5 & $21.16_{\pm 3.31}$ & $27.00_{\pm 5.59}$ & $27.70_{\pm 4.77}$ & $20.91_{\pm 4.31}$ & $11.17_{\pm 2.56}$ & $18.95_{\pm 3.97}$ \\
    Phi3-vision-128k & $2.05_{\pm 1.17}$ & $0.00$ & $0.00$ & $1.17_{\pm 1.14}$ & $0.70_{\pm 0.73}$ & $2.05_{\pm 1.36}$ \\
    MiniGPT4-7b & $1.35_{\pm 0.96}$ & $2.95_{\pm 2.10}$ & $3.41_{\pm 1.93}$ & $3.43_{\pm 1.89}$ & $2.05_{\pm 1.20}$ & $1.94_{\pm 1.36}$ \\
    mPLUG-Owl-2 & $0.00$ & $0.00$ & $0.00$ & $0.00$ & $0.00$ & $0.00$ \\
    LRV-MiniGPT4-7b & $2.81_{\pm 1.43}$ & $1.49_{\pm 1.47}$ & $1.15_{\pm 1.17}$ & $1.10_{\pm 1.13}$ & $1.41_{\pm 1.00}$ & $2.01_{\pm 1.40}$ \\
    InternVL1.5 & $21.16_{\pm 3.31}$ & $27.00_{\pm 5.59}$ & $27.70_{\pm 4.77}$ & $20.91_{\pm 4.31}$ & $11.17_{\pm 2.56}$ & $18.95_{\pm 3.97}$ \\
    LLaVa-1.5-7b & $0.00$ & $0.00$ & $0.00$ & $0.00$ & $0.00$ & $0.00$ \\
    Qwen2vl-instruct-7b  & $1.40_{\pm 1.00}$ & $1.52_{\pm 1.51}$ & $0.00$ & $0.00$ & $2.10_{\pm 1.21}$ & $3.84_{\pm 1.97}$ \\
    \bottomrule
    \end{tabular}
}
\label{tab:multiple_choice_object_categories}
\end{table*}

\begin{table*}[ht]
\centering
\caption{Attribute category performance breakdown for multiple choice QA. We report mean $\pm$ standard deviation. Results demonstrate how models handle different descriptive properties, from concrete visual attributes (colors, materials) to more abstract characteristics (environmental conditions, physical properties). Notable performance gaps emerge between attribute categories, with colors generally being easier to detect than environmental descriptors. }
\resizebox{0.85\textwidth}{!}{
    \begin{tabular}{lcccccc}
    \toprule
    \textbf{Model} & \textbf{Colors} & \textbf{Numbers} & \textbf{Materials} & \textbf{Physical} & \textbf{Environmental} & \textbf{Other} \\
    \midrule
    GPT 5 & $95.49_{\pm 1.24}$ & $75.70_{\pm 2.95}$ & $93.06_{\pm 3.38}$ & $84.04_{\pm 8.09}$ & $57.88_{\pm 14.19}$ & $73.22_{\pm 8.41}$ \\
    GPT 4.1 Mini & $31.59_{\pm 2.62}$ & $6.87_{\pm 1.75}$ & $11.82_{\pm 4.18}$ & $5.06_{\pm 5.31}$ & $16.74_{\pm 10.83}$ & $10.12_{\pm 5.49}$ \\
    Gemini 2.5 Pro & $93.50_{\pm 1.37}$ & $85.83_{\pm 2.45}$ & $91.24_{\pm 3.67}$ & $89.21_{\pm 7.22}$ & $66.58_{\pm 13.63}$ & $73.34_{\pm 7.89}$ \\
    Gemini 2.5 Flash Lite & $9.42_{\pm 1.66}$ & $4.42_{\pm 1.47}$ & $6.78_{\pm 3.24}$ & $20.96_{\pm 9.44}$ & $8.63_{\pm 7.99}$ & $10.15_{\pm 5.57}$ \\
    \midrule
    InstructBlip-T5xxl & $61.35_{\pm 2.84}$ & $56.73_{\pm 3.46}$ & $63.93_{\pm 6.32}$ & $68.03_{\pm 10.83}$ & $48.87_{\pm 14.95}$ & $47.11_{\pm 8.87}$ \\
    InternVL1.5 & $16.40_{\pm 2.13}$ & $3.94_{\pm 1.44}$ & $15.60_{\pm 4.73}$ & $20.59_{\pm 9.24}$ & $0.00$ & $16.35_{\pm 7.05}$ \\
    Phi3-vision-128k & $1.57_{\pm 0.72}$ & $0.00$ & $0.00$ & $0.00$ & $0.00$ & $3.31_{\pm 3.29}$ \\
    MiniGPT4-7b & $2.58_{\pm 0.89}$ & $2.42_{\pm 1.04}$ & $1.76_{\pm 1.71}$ & $0.00$ & $0.00$ & $3.28_{\pm 3.27}$ \\
    mPLUG-Owl-2 & $0.00$ & $0.00$ & $0.00$ & $0.00$ & $0.00$ & $0.00$ \\
    LRV-MiniGPT4-7b & $2.27_{\pm 0.87}$ & $2.96_{\pm 1.16}$ & $1.66_{\pm 1.72}$ & $0.00$ & $0.00$ & $6.59_{\pm 4.36}$ \\
    LLaVa-1.5-7b & $0.00$ & $0.00$ & $0.00$ & $0.00$ & $0.00$ & $0.00$ \\
    Qwen2vl-instruct-7b  & $0.97_{\pm 0.55}$ & $0.48_{\pm 0.49}$ & $0.00$ & $0.00$ & $0.00$ & $0.00$ \\
    \bottomrule
    \end{tabular}
}
\label{tab:multiple_choice_attribute_categories}
\end{table*}

\subsection{Open-ended question answering}

Unlike multiple-choice tasks, where models select from given options, open-ended questions require models to formulate their own responses. This format more closely mirrors real-world deployment scenarios where models must generate explanations or decisions without explicit guidance about potential conflicts.

We evaluate whether models can naturally identify contradictions between visual and textual inputs, or whether they default to following one modality while ignoring the other. The evaluation protocol categorizes responses into four classes: correctly identifying \textbf{conflicts}, following \textbf{image}-based information, adhering to \textbf{text}-based descriptions, or producing \textbf{incorrect}/unintelligible responses.

\paragraph{Overall performance on open-ended tasks.}

This section presents a comprehensive analysis of open-ended conflict detection performance, summarizing key results from the main paper. Table~\ref{tab:open_ended_modality_preference} provides the complete breakdown of response categories, while Table~\ref{tab:open_ended_scores} and Figure~\ref{fig:modality_preference_open_ended} in the main text focus on conflict detection rates and modality bias patterns separately.

\begin{table*}[ht!]
\centering
\caption{Evaluation of models on open-ended QA using relaxed string matching. Columns indicate the percentage of predictions matching Conflict, Image or Text, with Incorrect denoting responses that match none of them. }
\resizebox{0.6\textwidth}{!}{
    \begin{tabular}{lcccc}
    \toprule
    \textbf{Model} & \textbf{Conflict} ($\uparrow$) & \textbf{Image} ($\downarrow$)& \textbf{Text} ($\downarrow$)& \textbf{Incorrect} ($\downarrow$)\\
    \midrule
    GPT 5 & $81.21_{\pm 1.08}$ & $1.11_{\pm 0.30}$ & $9.49_{\pm 0.81}$ & $8.21_{\pm 0.78}$ \\
    GPT 4.1 Mini & $40.03_{\pm 1.37}$ & $40.39_{\pm 1.35}$ & $19.59_{\pm 1.09}$ & $5.53_{\pm 0.61}$ \\
    Gemini 2.5 Pro & $91.61_{\pm 0.76}$ & $0.84_{\pm 0.25}$ & $7.78_{\pm 0.70}$ & $0.30_{\pm 0.15}$ \\
    Gemini 2.5 Flash Lite & $20.94_{\pm 1.12}$ & $49.73_{\pm 1.35}$ & $21.64_{\pm 1.18}$ & $9.39_{\pm 0.82}$ \\
    \midrule
    InternVL1.5 & $18.72_{\pm 1.11}$ & $22.67_{\pm 1.19}$ & $55.51_{\pm 1.42}$ & $6.31_{\pm 0.70}$ \\
    mPLUG-Owl-1 & $6.57_{\pm 0.69}$ & $15.91_{\pm 1.03}$ & $24.17_{\pm 1.22}$ & $61.50_{\pm 1.38}$ \\
    mPLUG-Owl-2 & $1.40_{\pm 0.33}$ & $1.02_{\pm 0.29}$ & $97.03_{\pm 0.48}$ & $0.93_{\pm 0.26}$ \\
    LLaVa-1.5-7b & $0.00$ & $34.67_{\pm 1.29}$ & $62.10_{\pm 1.39}$ & $4.97_{\pm 0.63}$ \\
    \bottomrule
    \end{tabular}
}
\label{tab:open_ended_modality_preference}
\end{table*}

\paragraph{LLM-as-judge.}
To assess the consistency and reliability of our relaxed string matching procedure for evaluating the open-ended task, we ran an LLM-as-a-judge using two models: Gemini 2.5 Pro and GPT-5. For each sample, we determined whether each model’s prediction matched one of the four categories: CONFLICT, IMAGE, TEXT, or NONE. We then recorded:

\begin{enumerate}
    \item \textbf{Gemini 2.5 Pro “Yes”} -- number of samples where Gemini assigned the category.
    \item \textbf{GPT-5 “Yes”} -- number of samples where GPT-5 assigned the category.
    \item \textbf{At least one “Yes”} -- number of samples where either model assigned the category.
    \item \textbf{Both “Yes”} -- number of samples where both models agreed on the category.
\end{enumerate}
This setup allows quantifying both individual model performance and inter-model agreement. Results are presented in Table~\ref{tab:llm_judge_evaluation}. The results reveal several trends, discussed below.

\begin{table*}[ht!]
\centering
\caption{LLM-as-judge evaluation of predictions from four evaluated open-ended models, using Gemini 2.5 Pro and GPT-5. Results show high consistency between judges and close alignment with string matching results (Table~\ref{tab:open_ended_modality_preference}).}
\resizebox{0.6\textwidth}{!}{
\begin{tabular}{l l c c c c}
    \toprule
    \textbf{Evaluated Model} & \textbf{Judge} & \textbf{Conflict} ($\uparrow$) & \textbf{Image} ($\downarrow$)& \textbf{Text} ($\downarrow$)& \textbf{Incorrect} ($\downarrow$) \\
    \midrule
    \multirow{4}{*}{InternVL1.5} & Gemini & $20.51_{\pm 1.08}$ & $23.83_{\pm 1.15}$ & $52.24_{\pm 1.37}$ & $3.43_{\pm 0.49}$ \\
    & GPT-5 & $20.54_{\pm 1.14}$ & $23.20_{\pm 1.21}$ & $52.04_{\pm 1.38}$ & $4.23_{\pm 0.56}$ \\
    & $\geq$1 & $21.33_{\pm 1.09}$ & $24.29_{\pm 1.18}$ & $52.30_{\pm 1.39}$ & $4.43_{\pm 0.58}$ \\
    & Both & $19.75_{\pm 1.14}$ & $22.81_{\pm 1.20}$ & $51.94_{\pm 1.42}$ & $3.19_{\pm 0.49}$ \\
    \midrule
    \multirow{4}{*}{mPLUG-Owl-1} & Gemini & $2.67_{\pm 0.45}$ & $14.24_{\pm 0.99}$ & $19.70_{\pm 1.11}$ & $61.85_{\pm 1.35}$ \\
    & GPT-5 & $2.43_{\pm 0.43}$ & $13.49_{\pm 0.94}$ & $19.23_{\pm 1.07}$ & $64.89_{\pm 1.29}$ \\
    & $\geq$1 & $3.79_{\pm 0.53}$ & $14.75_{\pm 0.99}$ & $20.51_{\pm 1.14}$ & $65.95_{\pm 1.40}$ \\
    & Both & $1.25_{\pm 0.31}$ & $12.97_{\pm 0.95}$ & $18.46_{\pm 1.11}$ & $60.79_{\pm 1.38}$ \\
    \midrule
    \multirow{4}{*}{mPLUG-Owl-2} & Gemini & $1.40_{\pm 0.33}$ & $0.78_{\pm 0.25}$ & $96.69_{\pm 0.50}$ & $1.08_{\pm 0.29}$ \\
    & GPT-5 & $1.41_{\pm 0.34}$ & $0.71_{\pm 0.24}$ & $96.79_{\pm 0.47}$ & $1.10_{\pm 0.30}$ \\
    & $\geq$1 & $1.38_{\pm 0.34}$ & $0.77_{\pm 0.24}$ & $97.04_{\pm 0.48}$ & $1.33_{\pm 0.32}$ \\
    & Both & $1.41_{\pm 0.33}$ & $0.71_{\pm 0.23}$ & $96.49_{\pm 0.50}$ & $0.85_{\pm 0.25}$ \\
     \midrule
    \multirow{4}{*}{LLaVa-1.5-7b} & Gemini & $0.24_{\pm 0.13}$ & $35.89_{\pm 1.32}$ & $60.31_{\pm 1.36}$ & $3.56_{\pm 0.51}$ \\
    & GPT-5 & $0.16_{\pm 0.11}$ & $36.11_{\pm 1.36}$ & $60.95_{\pm 1.34}$ & $2.70_{\pm 0.45}$ \\
    & $\geq$1 & $0.31_{\pm 0.16}$ & $36.22_{\pm 1.38}$ & $61.17_{\pm 1.36}$ & $3.66_{\pm 0.53}$ \\
    & Both & $0.08_{\pm 0.08}$ & $35.76_{\pm 1.35}$ & $60.20_{\pm 1.32}$ & $2.64_{\pm 0.46}$ \\
    \bottomrule
\end{tabular}
}
\label{tab:llm_judge_evaluation}
\end{table*}

\textit{High consistency across evaluation methods.}  The comparison between the string matching evaluation in Table~\ref{tab:open_ended_modality_preference} and LLM-as-judge evaluation in Table~\ref{tab:llm_judge_evaluation} revels remarkably consistent results, demonstrating the reliability of both approaches. Across all models, the differences between string matching and LLM-based evaluation are minimal, typically within 1-3 percentage points. For instance, InternVL 1.5's conflict detection accuracy shows only a 1.79 point difference (18.72\% string matching vs. 20.51\% Gemini), while modality bias patterns remain nearly identical. 

\textit{Inter-judge agreement and reliability.} The strong agreement between Gemini 2.5 Pro and GPT-5 as judges (differences $<1\%$  across most metrics) validates the robustness of LLM-based evaluation. This consistency suggests that both judge models apply similar semantic understanding when categorizing responses, reducing concerns about judge-specific biases.

These findings strengthen confidence in our evaluation methodology and suggest that either approach can reliably assess model performance on \bench.

\paragraph{Performance across object and attribute categories on open-ended tasks.}
We further analyze category-specific performance in the open-ended setting to understand how different types of contradictions affect free-form reasoning capabilities. This breakdown across object and attribute categories reveals whether the patterns observed in multiple-choice evaluation persist when models must generate their own explanations rather than select from predefined options. Tables~\ref{tab:open_ended_object_categories}
and~\ref{tab:open_ended_attribute_categories} performance of several open source and closed source models on the open-ended task across object and attribute categories. The evaluation is performed using relaxed string matching.

\begin{table*}[ht!]
\centering
\caption{Object category performance of several open source and closed source models on the open-ended task using relaxed string matching evaluation. We report mean $\pm$ standard deviation.}
\resizebox{0.8\textwidth}{!}{
    \begin{tabular}{lcccccc}
    \toprule
    \textbf{Model} & \textbf{Animals} & \textbf{Vehicles} & \textbf{Food} & \textbf{Sports} & \textbf{Household} & \textbf{Other} \\
    \midrule
    GPT 5 & $96.55_{\pm 1.50}$ & $89.56_{\pm 3.74}$ & $74.73_{\pm 4.55}$ & $86.82_{\pm 3.47}$ & $64.48_{\pm 4.12}$ & $67.53_{\pm 4.42}$ \\
    GPT 4.1 Mini & $54.82_{\pm 4.21}$ & $43.27_{\pm 6.16}$ & $38.86_{\pm 5.29}$ & $38.13_{\pm 4.92}$ & $33.52_{\pm 3.98}$ & $31.38_{\pm 4.53}$ \\
    Gemini 2.5 Pro & $98.67_{\pm 0.93}$ & $95.58_{\pm 2.39}$ & $92.99_{\pm 2.68}$ & $95.56_{\pm 2.12}$ & $84.72_{\pm 3.00}$ & $82.34_{\pm 3.82}$ \\
    Gemini 2.5 Flash Lite & $21.24_{\pm 3.41}$ & $13.46_{\pm 4.17}$ & $11.34_{\pm 3.33}$ & $8.67_{\pm 2.93}$ & $16.66_{\pm 3.04}$ & $20.63_{\pm 4.08}$ \\
    \midrule
    InternVL1.5 & $26.10_{\pm 3.63}$ & $31.37_{\pm 5.59}$ & $21.71_{\pm 4.17}$ & $31.54_{\pm 4.55}$ & $14.64_{\pm 2.93}$ & $20.56_{\pm 3.96}$ \\
    mPLUG-Owl-1 & $4.77_{\pm 1.77}$ & $7.36_{\pm 3.01}$ & $10.31_{\pm 3.22}$ & $3.27_{\pm 1.83}$ & $5.55_{\pm 1.94}$ & $4.91_{\pm 2.19}$ \\
    mPLUG-Owl-2 & $0.00$ & $0.00$ & $2.38_{\pm 1.63}$ & $1.02_{\pm 1.04}$ & $0.70_{\pm 0.69}$ & $2.91_{\pm 1.69}$ \\
    LLaVa-1.5-7b & $0.00$ & $0.00$ & $0.00$ & $0.00$ & $0.00$ & $0.00$ \\
    \bottomrule
    \end{tabular}
}
\label{tab:open_ended_object_categories}
\end{table*}

\begin{table*}[ht!]
\centering
\caption{Attribute category performance of several open source and closed source models on the open-ended task using relaxed string matching evaluation. We report mean $\pm$ standard deviation.}
\resizebox{0.8\textwidth}{!}{
    \begin{tabular}{lcccccc}
    \toprule
    \textbf{Model} & \textbf{Colors} & \textbf{Numbers} & \textbf{Materials} & \textbf{Physical} & \textbf{Environmental} & \textbf{Other} \\
    \midrule
    GPT 5 & $93.52_{\pm 1.41}$ & $67.33_{\pm 3.39}$ & $91.24_{\pm 3.81}$ & $68.17_{\pm 10.46}$ & $49.61_{\pm 12.07}$ & $76.63_{\pm 8.31}$ \\
    GPT 4.1 Mini & $66.18_{\pm 2.75}$ & $3.36_{\pm 1.19}$ & $27.33_{\pm 6.00}$ & $5.45_{\pm 5.21}$ & $31.73_{\pm 11.54}$ & $50.01_{\pm 9.71}$ \\
    Gemini 2.5 Pro & $96.48_{\pm 1.08}$ & $86.50_{\pm 2.34}$ & $96.60_{\pm 2.37}$ & $89.16_{\pm 7.11}$ & $62.86_{\pm 12.15}$ & $80.85_{\pm 7.62}$ \\
    Gemini 2.5 Flash Lite & $34.21_{\pm 2.65}$ & $16.10_{\pm 2.58}$ & $22.24_{\pm 5.63}$ & $16.05_{\pm 8.49}$ & $12.22_{\pm 8.40}$ & $26.83_{\pm 8.70}$ \\
    \midrule
    InternVL1.5 & $16.80_{\pm 2.06}$ & $13.15_{\pm 2.39}$ & $10.32_{\pm 3.87}$ & $0.00$ & $18.91_{\pm 9.80}$ & $7.74_{\pm 5.07}$ \\
    mPLUG-Owl-1 & $7.79_{\pm 1.47}$ & $5.81_{\pm 1.60}$ & $6.87_{\pm 3.29}$ & $15.73_{\pm 8.64}$ & $0.00$ & $3.87_{\pm 3.87}$ \\
    mPLUG-Owl-2 & $0.31_{\pm 0.31}$ & $1.46_{\pm 0.85}$ & $10.09_{\pm 3.89}$ & $5.17_{\pm 5.13}$ & $0.00$ & $0.00$ \\
    LLaVa-1.5-7b & $0.00$ & $0.00$ & $0.00$ & $0.00$ & $0.00$ & $0.00$ \\
    \bottomrule
    \end{tabular}
}
\label{tab:open_ended_attribute_categories}
\end{table*}

\subsection{Finetuning}
\label{app:sec:fine_tuning}


While our finetuning demonstrates clear improvements on conflict detection in \S\ref{subsec:lora_finetuning_main}, we evaluate performance on standard vision-language benchmarks to understand the broader impact of task-specific adaptation. We emphasize that our training data is purposefully constructed for the conflict detection task and does not aim to improve general vision-language capabilities. POPE \citep{li2023evaluating} tests object hallucination via yes/no questions about object presence across random, popular, and adversarial settings, totaling 9$k$ samples. 
OKVQA~\citep{marino2019ok} and  GQA~\citep{hudson2019gqa} (5$k$ and 12$k$ samples)
measure general visual question answering capabilities requiring external knowledge and spatial reasoning, respectively.

Tables~\ref{tab:pope} and \ref{tab:okvqa_gqa} reveal consistent patterns across benchmarks. mPLUG-Owl-1, which had poor baseline performance, shows consistent improvements across nearly all metrics after finetuning. In contrast, for LLaVa-1.5-7b, finetuning leads to substantial drops on most tasks. This indicates that the specialized training serves as general capability enhancement for initially weak models, while potentially causing capability regression in stronger baseline models.  We note that benchmarks like OKVQA and GQA provide their own training sets that could be used for finetuning -- for building a general-purpose expert model, we would recommend finetuning on a mixture of diverse datasets rather than solely on \bench.


These results serve as a reminder that task-specific finetuning, while effective for the target task, may require additional considerations -- such as multi-task training or regularization strategies -- to maintain performance on out-of-distribution evaluation scenarios during deployment.


 \begin{table}[ht]
    \centering
    \caption{Accuracy (\%) on POPE-COCO across random/popular/adversarial settings. We report mean $\pm$ standard deviation computed via bootstrap resampling with 1000 iterations.}
    \resizebox{0.95\columnwidth}{!}{
        \begin{tabular}{llll}
        \toprule
        \textbf{Model} & \textbf{Random} & \textbf{Popular} & \textbf{Adversarial}  \\
        \midrule
        LLaVa-1.5-7b &  $89.60_{\pm 0.31}$ & $86.22_{\pm 0.40}$   & $79.68_{\pm 0.50}$   \\
        LLaVa-1.5-7b-ft &  $83.42_{\pm 0.44}$ & $84.06_{\pm 0.49}$ & $81.02_{\pm 0.55}$  $_\uparrow$\\
        \midrule
        mPLUG-Owl-1 &  $35.32_{\pm 0.73}$  & $31.98_{\pm 0.71}$ & $32.28_{\pm 0.72}$ \\
        mPLUG-Owl-1-ft & $53.07_{\pm 0.81}$ $_\uparrow$ & $46.91_{\pm 0.84}$ $_\uparrow$  &  $45.37_{\pm 0.85}$$_\uparrow$ \\
        \bottomrule
        \end{tabular}
    }
    \label{tab:pope}
\end{table}

 \begin{table}[ht]
    \centering
    \caption{Performance on OKVQA and GQA. 
    }
    \resizebox{0.75\columnwidth}{!}{
        \begin{tabular}{lll}
        \toprule
        \textbf{Model} & \textbf{OKVQA} & \textbf{GQA} \\
        \midrule
        LLaVa-1.5-7b & $60.84_{\pm 0.56}$  &  $58.96_{\pm 0.21}$   \\
        LLaVa-1.5-7b-ft & $43.93_{\pm 0.54}$  & $44.58_{\pm 0.20}$   \\
        \midrule
        mPLUG-Owl-1 &  $35.39_{\pm 0.54}$ &  $28.05_{\pm 0.17}$ \\
        mPLUG-Owl-1-ft & $44.18_{\pm 0.55}$  $_\uparrow$  & $35.06_{\pm 0.18}$   $_\uparrow$ \\
        \bottomrule
        \end{tabular}
    }
    \label{tab:okvqa_gqa}
\end{table}


\subsection{Spatial conflicts}

The open-ended evaluation on spatial conflicts presented in Table~\ref{tab:open_ended_results_spatial} reveals similar patterns to the multiple-choice results in App~\ref{app:subsec:multiple_choice}, with even more pronounced degradation. GPT-5 conflict detection drops from 81.21\% on object and attribute conflicts to 56.20\% on spatial relations. InternVL1.5 also shows a decrease in conflict detection (18.72\% to 9.58\%). InstructBlip-T5xxl achieves near-zero conflict detection (0.36\%), predominantly producing text-based answers (58.98\%) or invalid outputs (34.38\%). These results demonstrate that spatial reasoning failures are consistent across evaluation formats.

\begin{table*}[th]
\centering
\footnotesize
\caption{Percentage of predictions matching the respective answer on the open ended task using \textbf{spatial} conflict types. The last column denotes cases where no match with any of the answers was found. The error bars$_{\pm}$ show standard deviation. 
}
\resizebox{0.6\textwidth}{!}{
    \begin{tabular}{l ccccc}
    \toprule
    \textbf{Model} & \textbf{Conflict} ($\uparrow$)& \textbf{Image} & \textbf{Text} & \textbf{Incorrect} \\
    \midrule
    GPT 5 & $56.20_{\pm 3.01}$ & $0.77_{\pm 0.54}$ & $11.34_{\pm 1.92}$ & $31.61_{\pm 2.91}$ \\
    \midrule
    InstructBlip-T5xxl & $0.36_{\pm 0.37}$ & $17.44_{\pm 2.39}$ & $58.98_{\pm 2.97}$ & $34.38_{\pm 2.81}$ \\
    InternVL1.5 & $9.58_{\pm 1.78}$ & $10.01_{\pm 1.79}$ & $48.41_{\pm 3.07}$ & $32.89_{\pm 2.76}$ \\
    \bottomrule
    \end{tabular}
}
\label{tab:open_ended_results_spatial}
\end{table*}


\section{Object Categories}
\label{app:sec:object_categories}

This section presents the object categories used in our evaluation framework. The categories are organized into four main domains: animals, transportation, food, sports, and household items. 

\subsection{Animals}
\begin{itemize}
    \item \textbf{Domestic/Farm Animals:} dog, cat, sheep, cow, cows, horse, horses, pig, goat, goats, cattle, donkeys, chicken, bull
    \item \textbf{Wild Animals:} elephant, zebra, zebras, giraffe, giraffes, bear, rhino, rhinoceros, rhinoceroses, rhinos, bird, birds, monkey, camel, antelope, deer, bison, wildebeest, hippopotamus, hippos, polar bear, brown bear, parrot, owl, crow, pigeon, butterfly, octopus, shark, fish, worm
    \item \textbf{Multiple/General:} elephants, dogs, cats, ducks, animals, kitten, puppy
\end{itemize}

\subsection{Transportation}
\begin{itemize}
    \item \textbf{Land Vehicles:} skateboard, skate board, bicycle, car, bus, scooter, motorcycle, train, truck, cars, tractor, automobile, motorcycles, bicycles, scooters, fire truck, police car, snowmobile, tow truck, pickup truck, train car, tour buses, bullet train
    \item \textbf{Air Vehicles:} plane, airplane, air plane, helicopter, fighter jet, commercial plane, fighter jets, commercial jets
    \item \textbf{Water Vehicles:} boat, boats, surf board, surfboard, surf boards, kayak, wakeboard
    \item \textbf{Transportation-Related:} bike, bikes, commercial jet, engine, trunk, trucks, road, tracks, track, windsail, parking meter
\end{itemize}

\subsection{Food}
\begin{itemize}
    \item \textbf{Fruits:} apples, bananas, banana, fruits, fruit, oranges, apple, pear, strawberries, blueberries, strawberry, banana peel, apple core
    \item \textbf{Vegetables:} vegetables, carrots, potatoes, broccoli, olives, tomatoes, carrot, cauliflower, onion rings, mushrooms, peppers, peas, spinach
    \item \textbf{Prepared Food:} pizza, burger, burgers, cake, hot dog, hot dogs, pastry, sandwich, donuts, cookies, food, noodles, hamburgers, hotdogs, cheese, sauce, sandwiches, quiche, meat, mead, beef, eggs, pasta, french fries, bread, hamburger, rice, cheeses, meats, fries, rice cake, cookie, pickle, piece of cake, slice of pizza, sausage
    \item \textbf{Drinks:} wine, beer, coffee, wine bottle, beer bottle, coffees, drinks, milk
    \item \textbf{Food Descriptors/Toppings:} toppings, sauces, greens, pepperoni
    \item \textbf{Food-Related Items:} bar b que, blender, bottle, bottles, bowls, chicken, dishes, fork, knife, olive, oysters, pears, pie, spoon, water, snails
\end{itemize}

\subsection{Sports}
\begin{itemize}
    \item \textbf{Sports Equipment:} ball, frisbee, snowboard, snow board, skis, ski, snowboards, baseball bat, baseball, tennis racket, tennis racquet, basketball, kite, kites, tennis ball, football, glove, bat, tennis, surfboard, surf board, surf boards, soccer, soccer ball, soccer balll, golf club, golf, golf ball, racket, racquet, shuttlecock, a snowboard, skateboard, skate board, bicycle
    \item \textbf{Sports Participants:} snowboarder, skier, skiier, batter, catcher, snowboarders, skiers, skateboarder, cyclist, surfer, kayaker, tennis players, basketball players, tennis player, basketball player, baseball player, football player, baseball players, football players, umpire, skateboarders, cyclists
    \item \textbf{Sports Venues:} tennis court, basketball court, skate park, ski lift
    \item \textbf{Sports-Related:} base ball, cricket, pitch, pitcher, ski lift, skis, surfer, track, wakeboard
\end{itemize}

\subsection{Household Items}
\begin{itemize}
    \item \textbf{Furniture:} chair, table, bed, sofa, couch, bench, coffee table, dining table, computer desk, nightstand, shelf, counter
    \item \textbf{Room Identifiers:} kitchen, bathroom, bedroom, living room, dining room
    \item \textbf{Bathroom Items:} toilet, sink, bathtub, shower, toothbrush, tooth brush, toilet tissue, soap, mirror, toilet bowl, shower curtain, towel rack, handicap bar
    \item \textbf{Kitchen Items:} fork, bowl, bowls, spoon, plate, plates, knife, knif, cup, trays, pots
    \item \textbf{Storage/Containers:} bag, trash can, laundry basket, baskets, mason jar, coffee cup, bottle, bottles, vases
    \item \textbf{Technology/Electronics:} phone, cell phone, laptop, laptops, keyboard, mouse, tablet, tablets, television, camera, phones, cellphones, wii, wii console, playstation, playstation console, xbox, remote, game remote, controller, refrigerator, oven, stove, microwave, dishwasher, washer, dryer, clothes washer, clothes dryer, blender, ice machine, coffee machine, printer, monitors, screen, clock, bell, ipod, microphones, speakers, equipment
    \item \textbf{Decor/Furnishing:} lamp, paintings, painting, picture frame, pillow, carpet, rug, sculptures, sculpture, statues, statutes, crosses, flags, flag
    \item \textbf{General Household:} furniture, window, door, towels, dishes, appliances, comb, rope, chain, scarf, mask, ties, scarves, backpack, coat, shirt, belt, hairbrush, aluminum foil, plastic wrap, stand, cart, books, book, glasses, handle, backrest, toys, toy, doll, accessories, clothing, swimsuit, dress, skirts, pants, roses, tulips, flowers, plants, leaves, branches, umbrella, umbrellas, hat, changing table, fire place, fireplace, pacifier, refrigerator magnet, urinal
\end{itemize}
\vspace{1em}

\section{Attribute Categories}
\label{app:sec:attribute_categories}

This section presents the comprehensive attribute categories used in our evaluation framework. The attributes are organized into five main domains: colors, numbers, materials, physical properties, and environmental conditions. 

\subsection{Colors}
\begin{itemize}
    \item \textbf{Single Colors:} blue, white, red, black, brown, green, yellow, orange, pink, grey, gray, purple, silver, tan, beige, cream, gold
    \item \textbf{Color Combinations:} black and white, blue and white, black and yellow, green and yellow, brown and white, black and red, black and gray, white and gray
    \item \textbf{Color Descriptors:} light blue, dark red, mint green, colorful, rainbow colored, monochrome, dark, light, color, colored, different colors, colors, browns, whites, rosy, colorfully, red-haired, blonde-haired, ginger, creamy
\end{itemize}

\subsection{Numbers}
\begin{itemize}
    \item \textbf{Basic Numbers:} one, two, three, four, five, six, seven
    \item \textbf{Written Numbers:} 2, 3, 25, 50
    \item \textbf{Ordinals:} first, second, third
    \item \textbf{Quantities:} a, another, solo, whole
    \item \textbf{Prices:} 11.98, 10.99
\end{itemize}

\subsection{Materials}
\begin{itemize}
    \item \textbf{Materials:} wooden, wood, metal, plastic, glass, ceramic, concrete, stainless steel, tile, brick, cement, marble, leather, fabric, steel, granite, stone, plywood, paper
    \item \textbf{Surface Qualities and Textures:} striped, polka dot, polka-dotted, polka dotted, tiled, plain, painted, polished, scratched, printed, stripped
\end{itemize}

\subsection{Physical Properties}
\begin{itemize}
    \item \textbf{Shapes:} square, round, circular, oval, rectangular, triangular
    \item \textbf{Physical Descriptors:} thick, thin, stuffed, sliced, ripe, unripe, wet, dry, clean, muddy, squares, wedges, opaque, clear, edge, back, duck shaped, fish shaped, horned, antlered
\end{itemize}

\subsection{Environmental Conditions}
\begin{itemize}
    \item \textbf{Weather:} sunny, snowy, cloudy, overcast, stormy, wet
    \item \textbf{Landscape/Terrain:} grassy, grass covered, snow covered, rocky, sandy, lush, dry, desert, tropical, remote, green, fenced
    \item \textbf{Water Depth:} knee deep, ankle deep
    \item \textbf{Light Conditions:} dim, bright
\end{itemize}

\vspace{1em}
\section{Qualitative examples}
\label{app:sec:qualitative_examples}

\begin{figure*}[ht!]
    \centering
    \begin{tabular}{ccc}
        \includegraphics[width=0.3\linewidth]{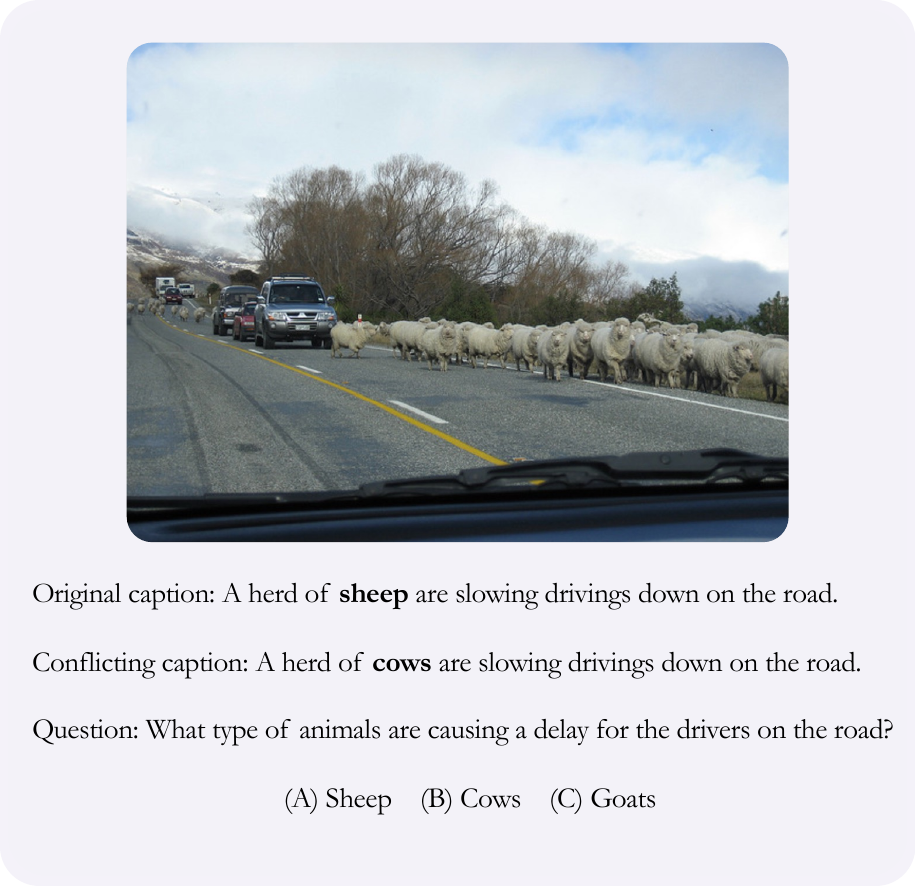} & 
        \includegraphics[width=0.3\linewidth]{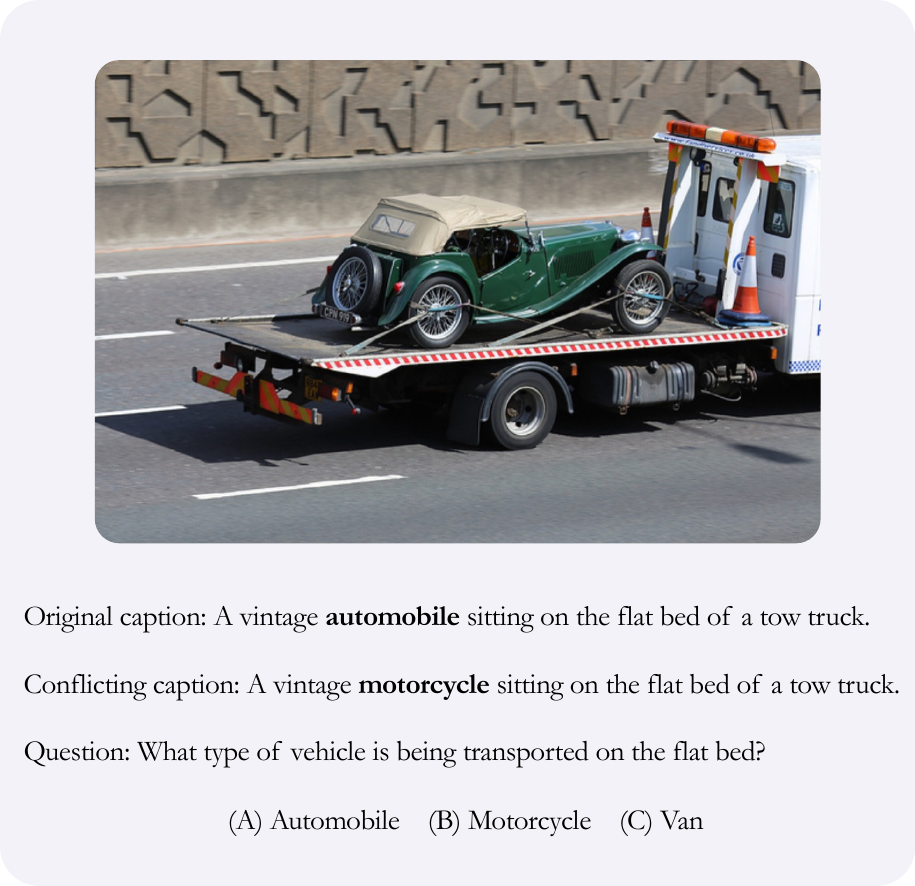} & 
        \includegraphics[width=0.3\linewidth]{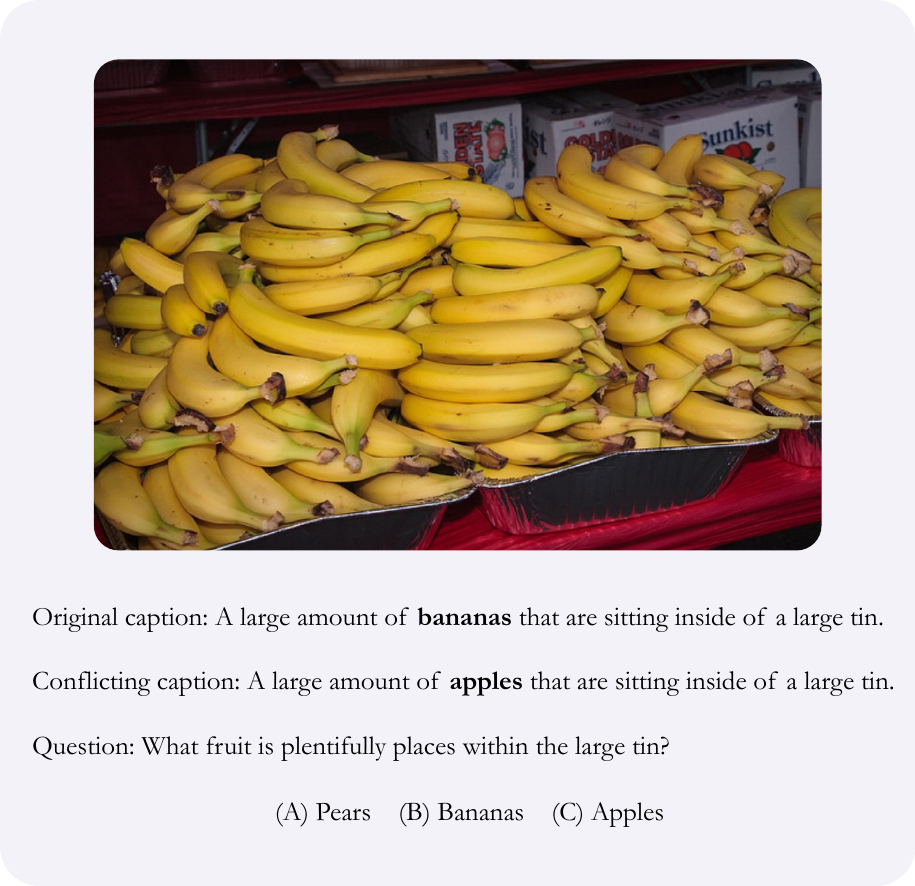} \\
        Animals & Vehicles & Food \\
        
        \includegraphics[width=0.3\linewidth]{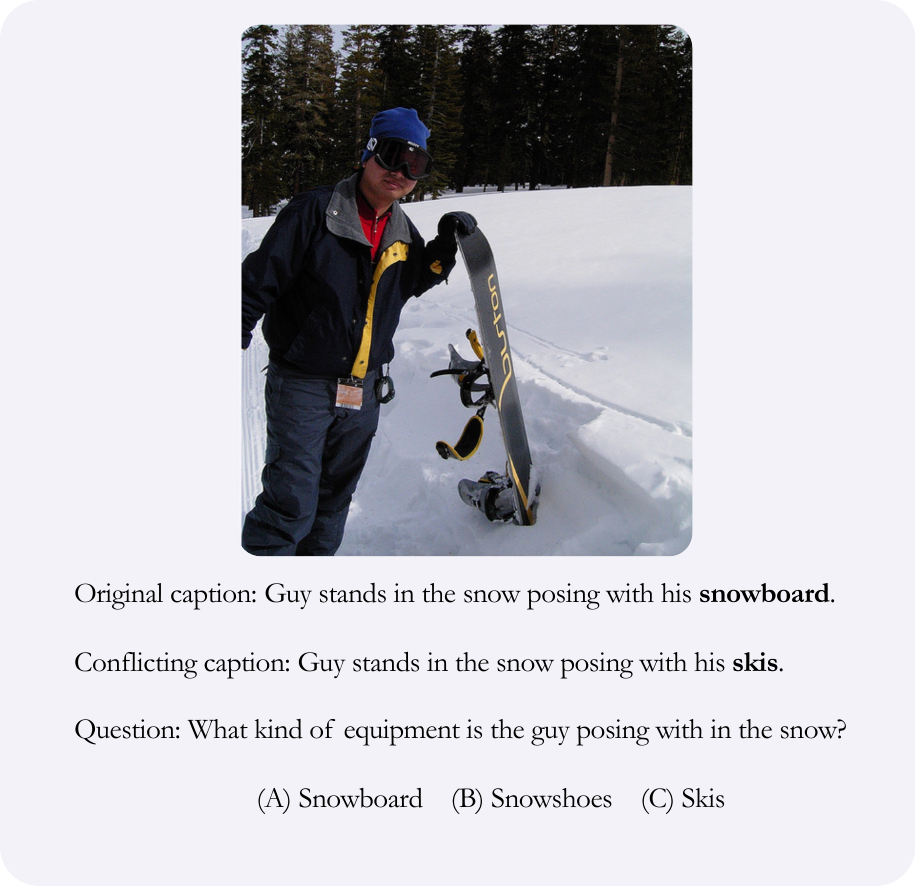} & 
        \includegraphics[width=0.3\linewidth]{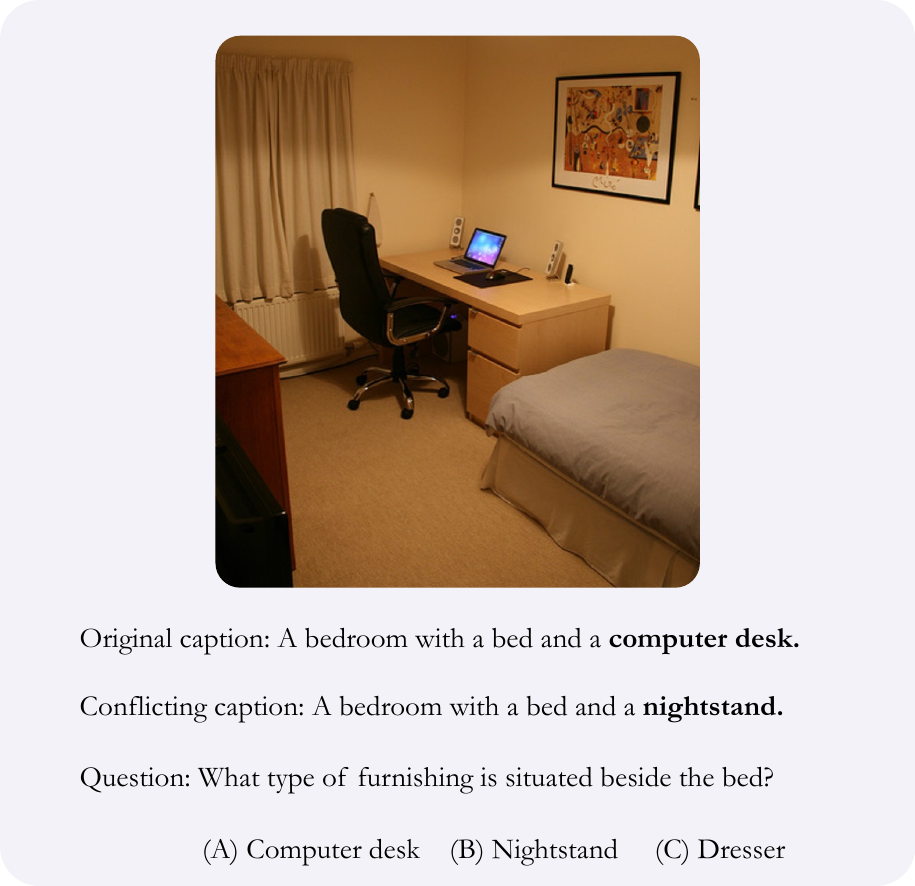} & 
        \includegraphics[width=0.3\linewidth]{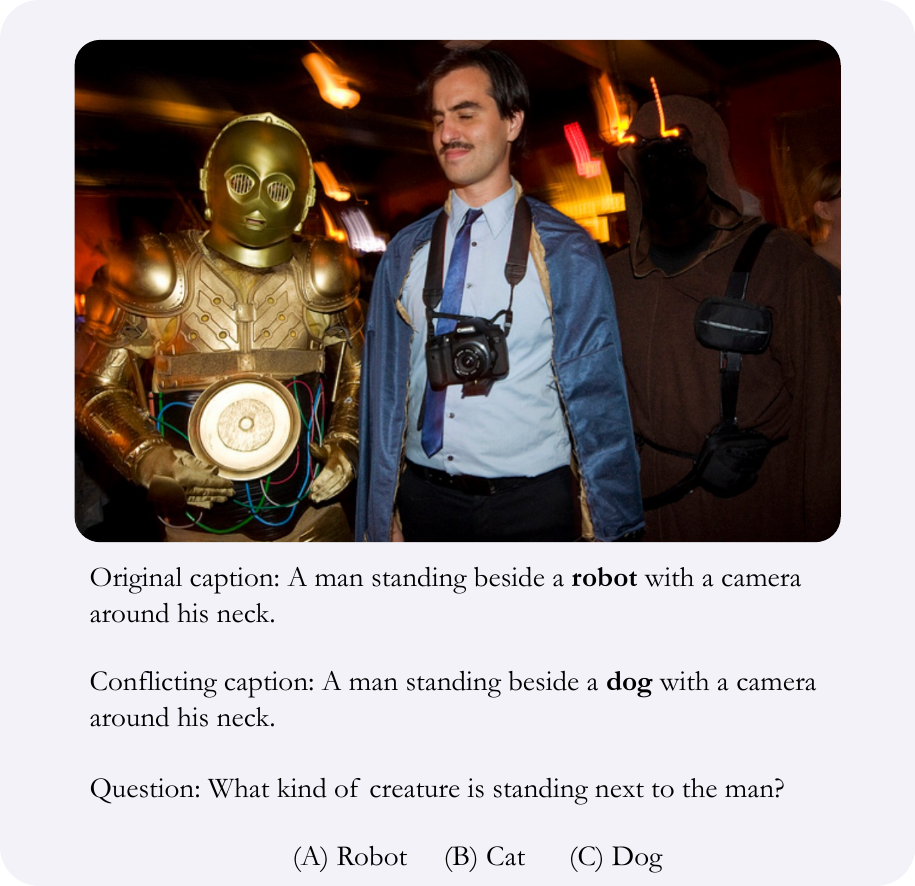} \\
        Sports & Household items & Other \\
        
        \includegraphics[width=0.3\linewidth]{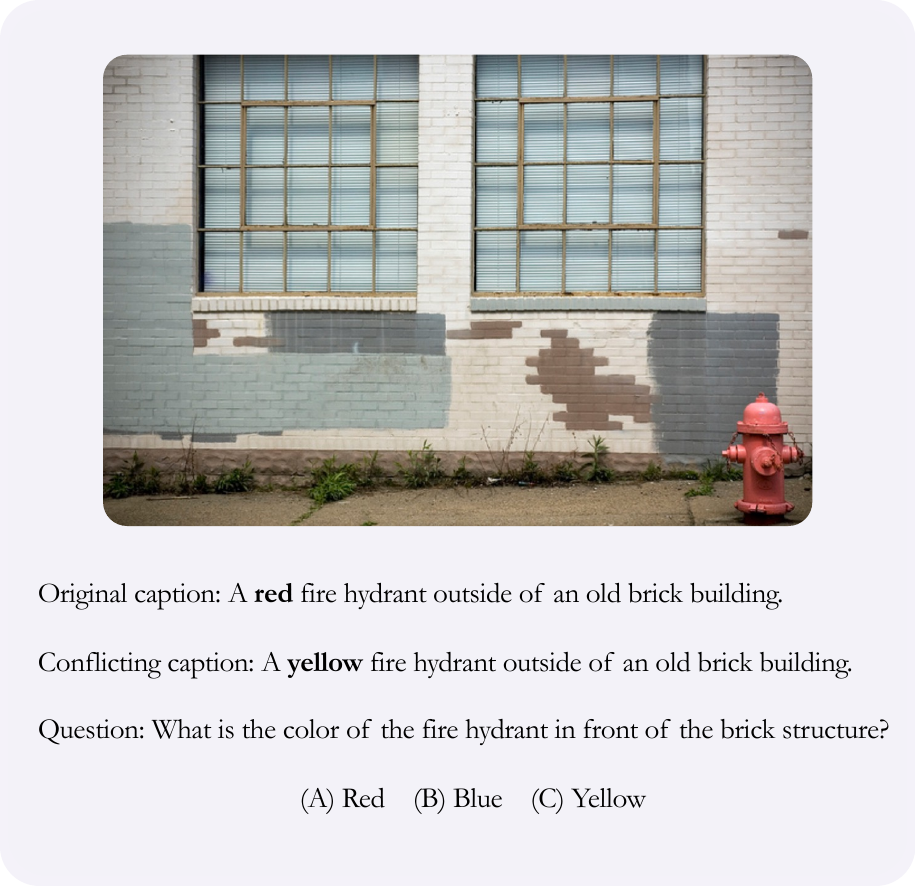} & 
        \includegraphics[width=0.3\linewidth]{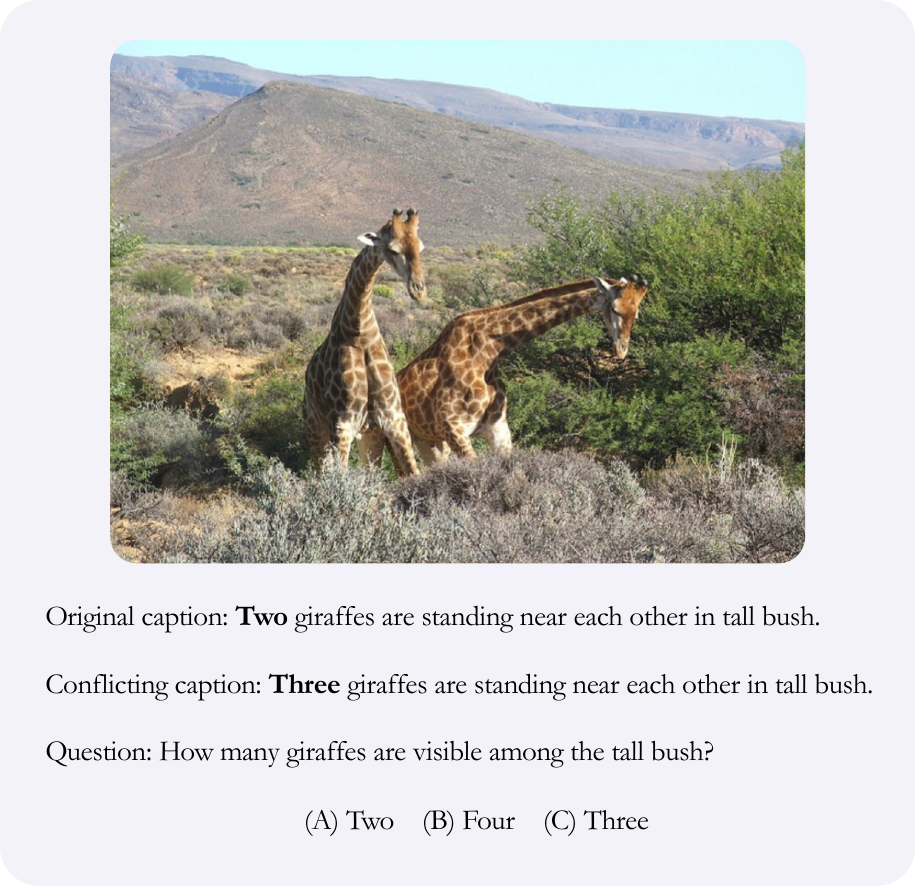} & 
        \includegraphics[width=0.3\linewidth]{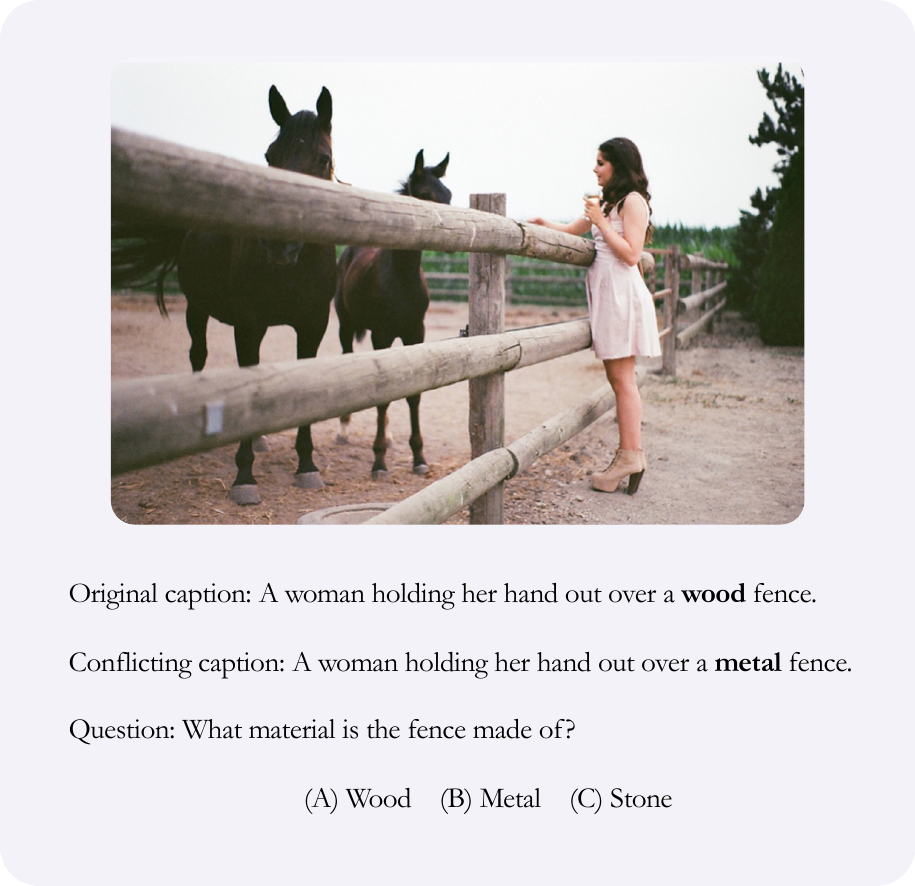} \\
        Colors & Numbers & Materials / Texture \\
        
        \includegraphics[width=0.3\linewidth]{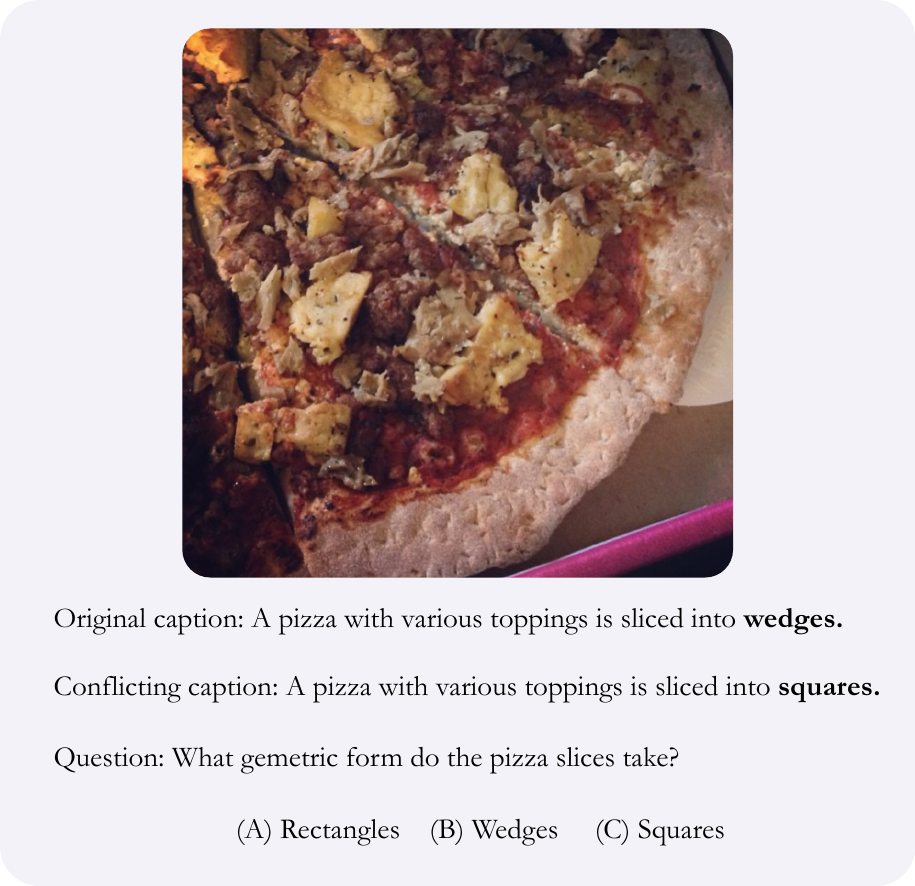} & 
        \includegraphics[width=0.3\linewidth]{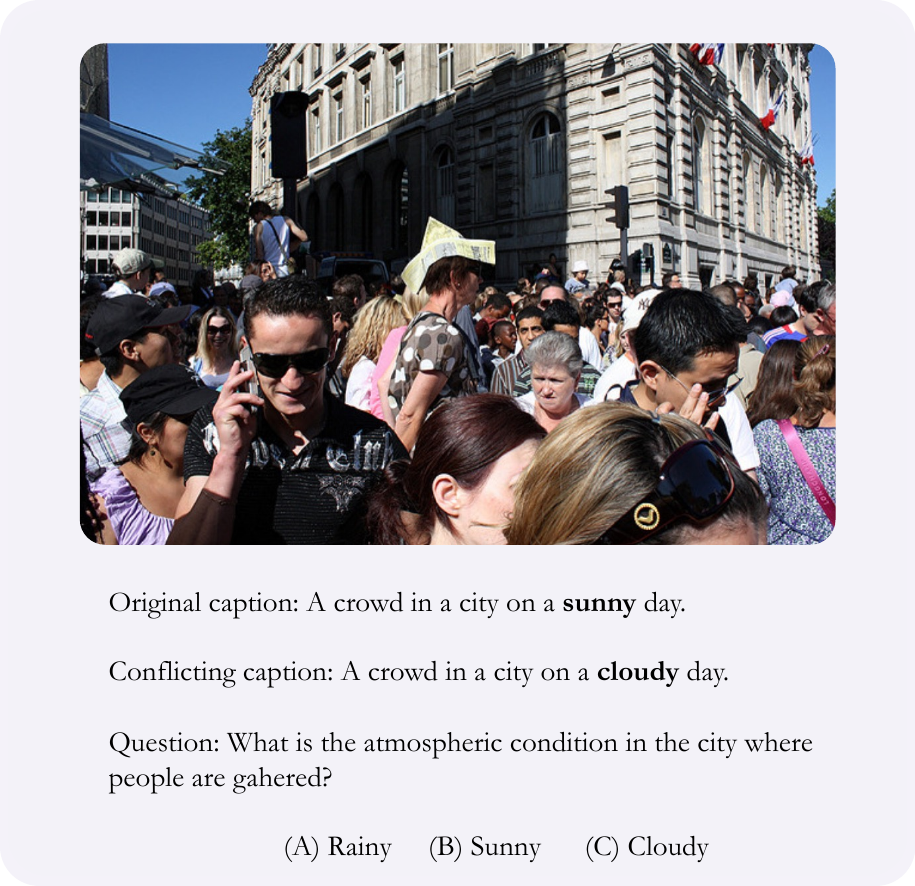} & 
        \includegraphics[width=0.3\linewidth]{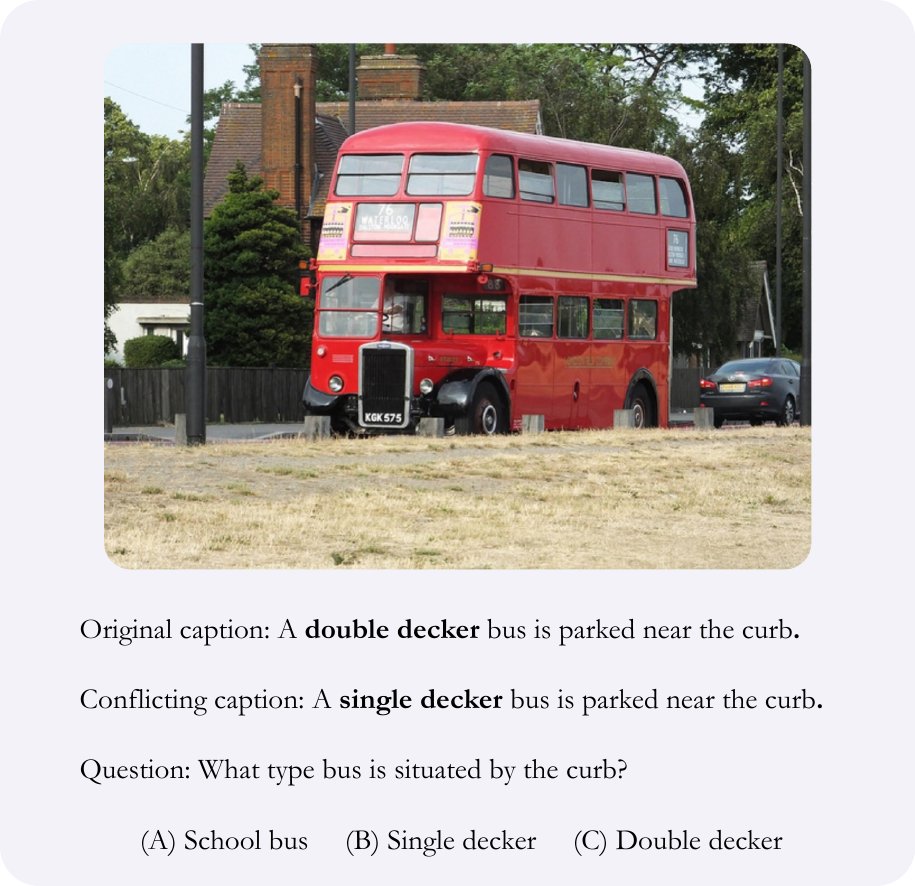} \\
        Physical properties & Environmental conditions & Other \\
    \end{tabular}
    \caption{Qualitative examples from \bench, illustrating the diversity of object and attribute categories. The dataset spans five object categories (top two rows): animals, transportation, food, sports, and household items, and five attribute categories (bottom two rows): colors, numbers, materials, physical properties, and environmental conditions.}
    \label{fig:qualitative_examples}
\end{figure*}

In this section, we present qualitative examples from \bench, illustrating the variety of object and attribute categories. The dataset covers five object categories -- animals, transportation, food, sports, and household items, and five attribute categories -- colors, numbers, materials, physical properties, and environmental conditions. Fig.~\ref{fig:qualitative_examples} shows qualitative examples from each category in \bench.

\section{Human validation}
\label{app:sec:human_validation}

Annotators assess three components for each sample: conflicting captions (verifying single-element modifications involving plausible, objective properties while avoiding impossible or subjective changes like man-to-woman), questions (confirming clarity, unambiguity, and focus on the changed element), and answers (checking for distinctiveness, objectivity, and visual observability while identifying problematic vague terms like "medium" or "beautiful"). Based on this assessment, annotators provide a single accept/reject decision for each sample. The exact instructions provided to the annotators are shown below, while Fig.~\ref{fig:human_validation_examples} depicts a few examples of accepted and rejected samples. Fig.~\ref{fig:examples_from_website} shows examples from the human verification interface. Annotators evaluate each sample using binary accept/reject votes to ensure the benchmark's reliability.

\begin{prompt}{Human validation instructions}
You will evaluate three things for each example:

\begin{enumerate}
    \item \textbf{Conflicting Caption}
    \begin{itemize}
        \item Did the caption change only one clear attribute or object (the change is marked in bold)?
        \item Is the change plausible and objective (e.g., color, number, shape, material, texture)?
    \end{itemize}

    \item \textbf{Question}
    \begin{itemize}
        \item Is the question clear and unambiguous?
        \item Does it focus on the changed attribute or object?
        \item Can it be answered using the given options?
    \end{itemize}

    \item \textbf{Answers}
    \begin{itemize}
        \item Are the answers distinct and not synonyms?
        \item Are they objective and visual (e.g., colors, numbers, objects)?
    \end{itemize}
\end{enumerate}

\noindent\textbf{Your task:} For each sample, mark whether it is sensible (Yes/No).
\end{prompt}

\begin{figure*}[ht!]
    \centering
    \begin{subfigure}[b]{0.32\textwidth}
        \centering
        \includegraphics[width=\textwidth]{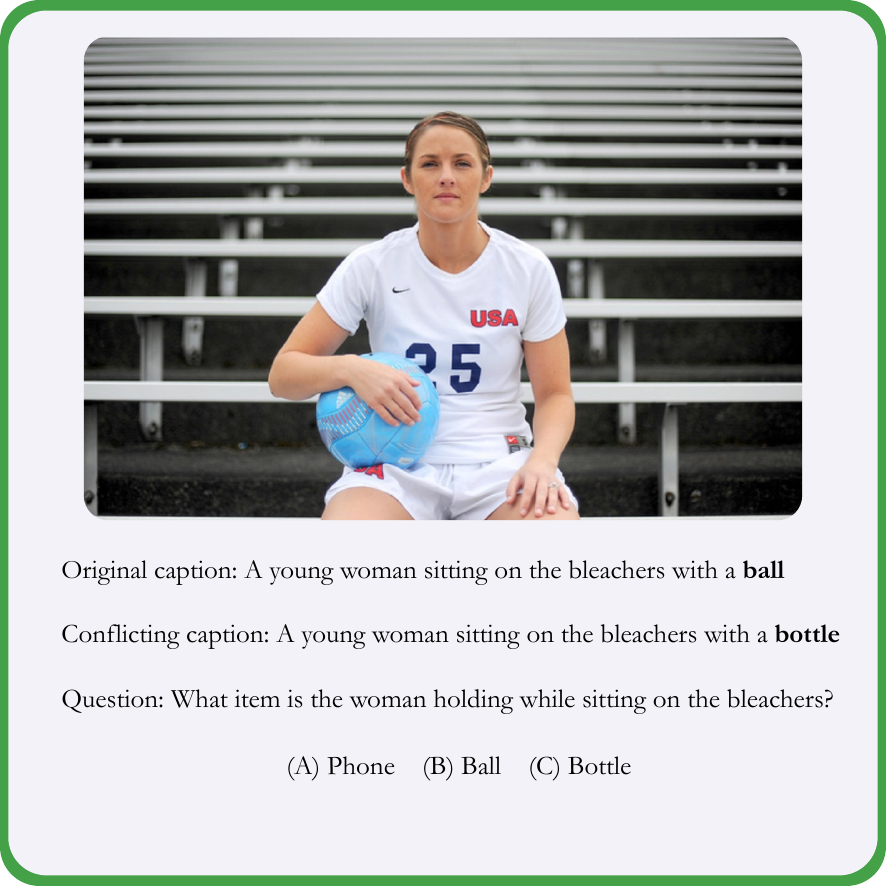}
        \caption{Changed object}
        \label{fig:sub1}
    \end{subfigure}
    \hfill
    \begin{subfigure}[b]{0.32\textwidth}
        \centering
        \includegraphics[width=\textwidth]{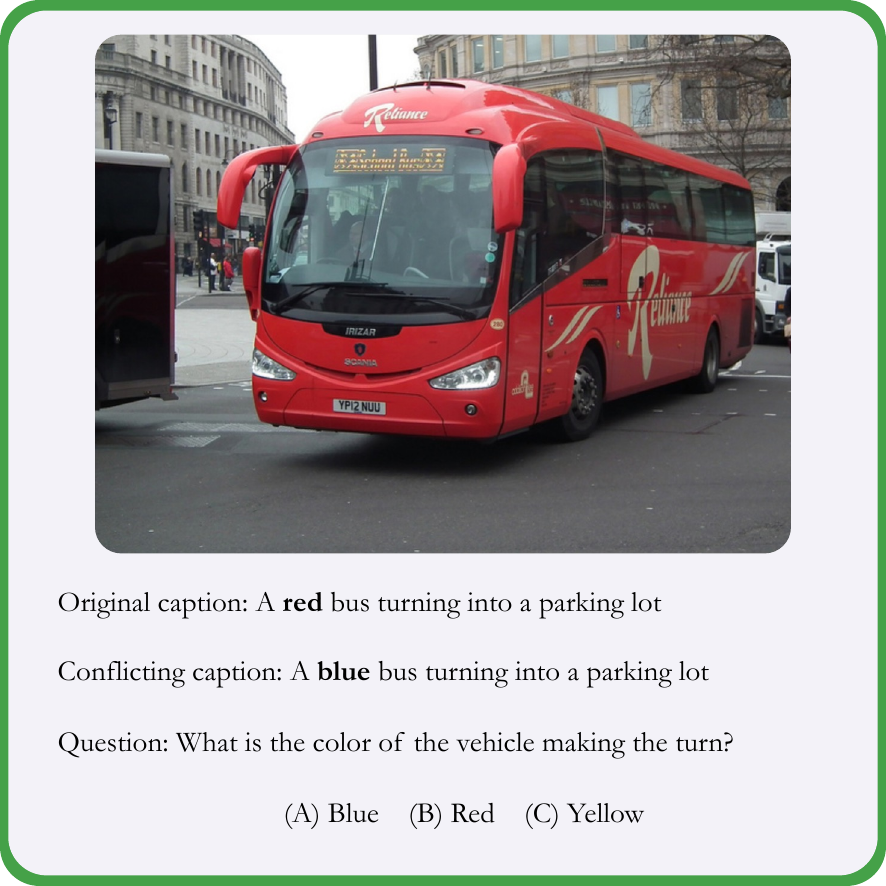}
        \caption{Changed attribute (color)}
        \label{fig:sub2}
    \end{subfigure}
    \hfill
    \begin{subfigure}[b]{0.32\textwidth}
        \centering
        \includegraphics[width=\textwidth]{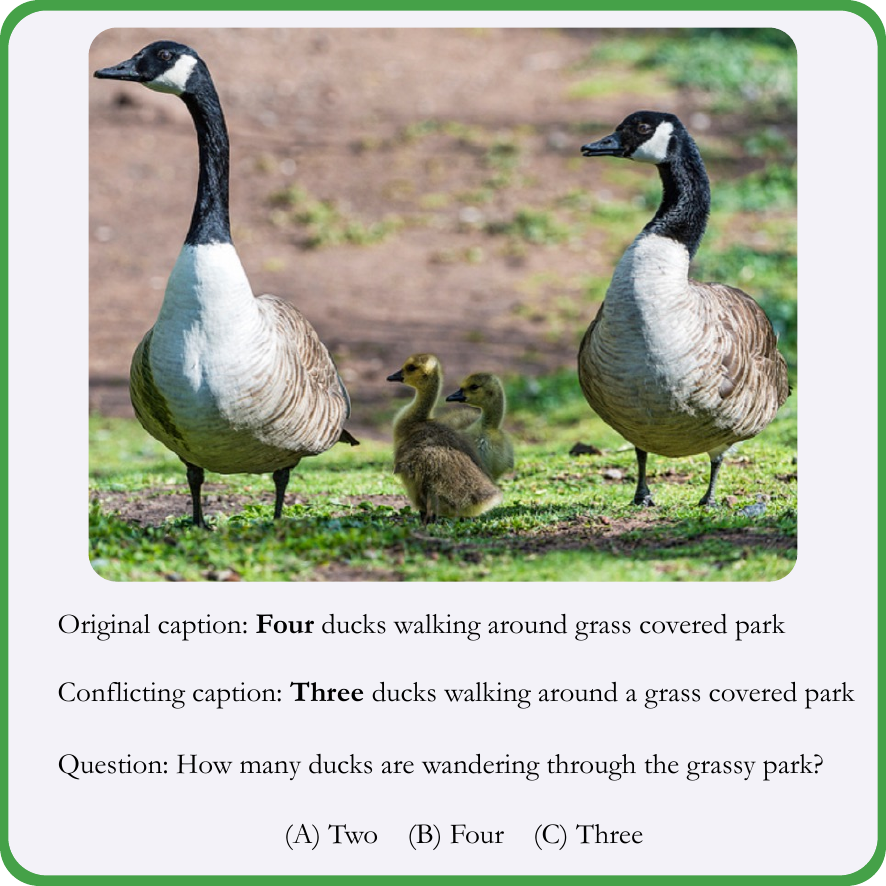}
        \caption{Changed attribute (number)}
        \label{fig:sub3}
    \end{subfigure}
    \vskip\baselineskip 
    \begin{subfigure}[b]{0.32\textwidth}
        \centering
        \includegraphics[width=\textwidth]{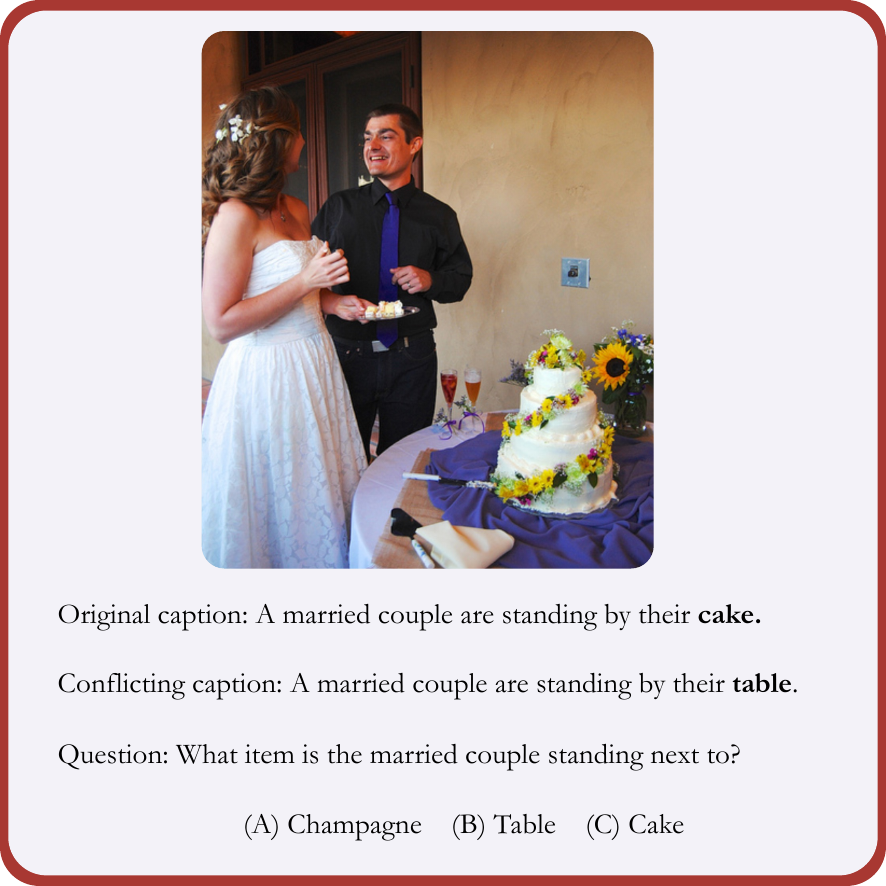}
        \caption{Problematic conflicting word}
        \label{fig:sub4}
    \end{subfigure}
    \hfill
    \begin{subfigure}[b]{0.32\textwidth}
        \centering
        \includegraphics[width=\textwidth]{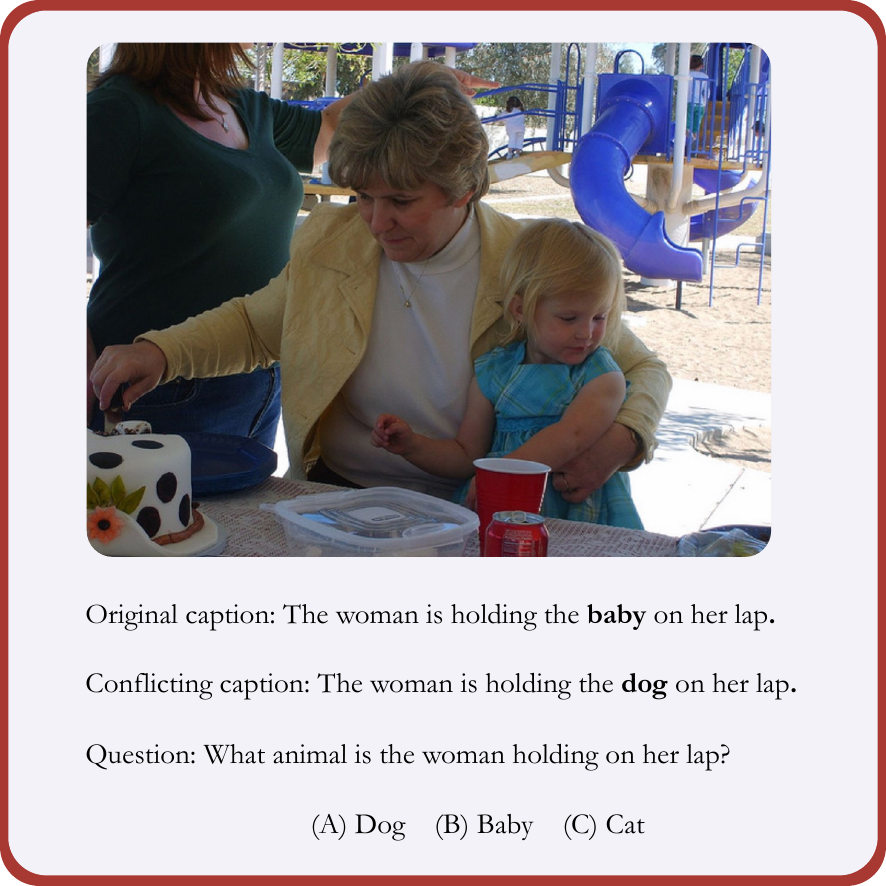}
        \caption{Problematic question}
        \label{fig:sub5}
    \end{subfigure}
    \hfill
    \begin{subfigure}[b]{0.32\textwidth}
        \centering
        \includegraphics[width=\textwidth]{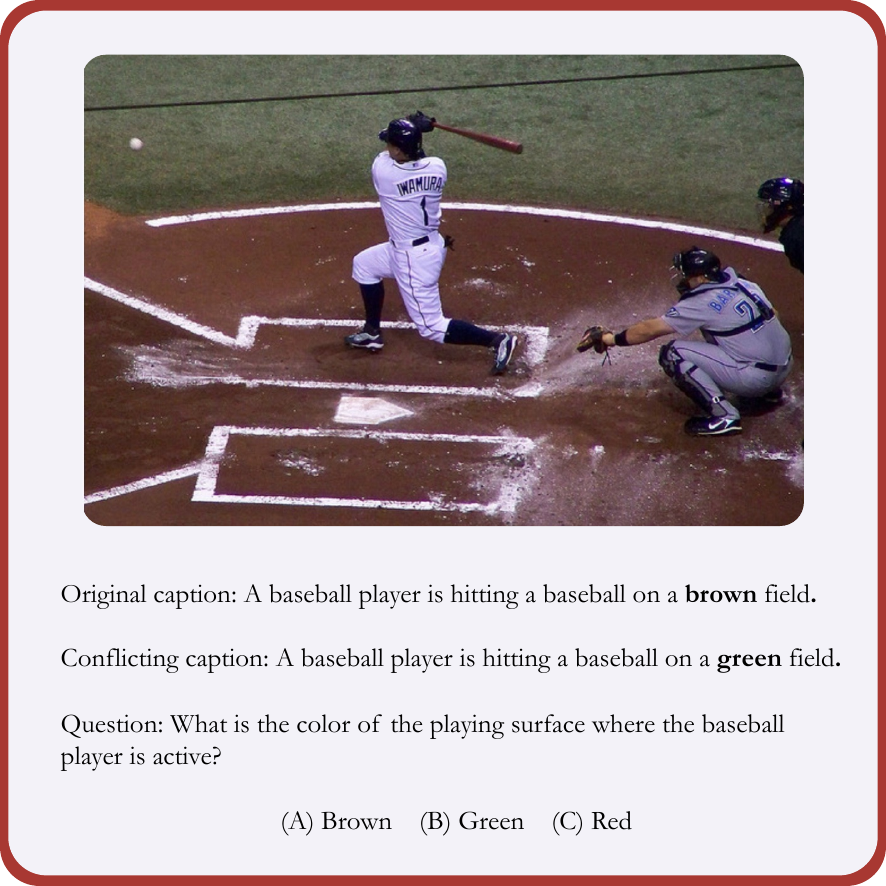}
        \caption{Problematic answers}
        \label{fig:sub6}
    \end{subfigure}
    \caption{Examples of accepted (\textbf{top row}) and rejected (\textbf{bottom row}) samples during human validation. Positive examples illustrate cases where the conflicting caption, question, and answers are clear and unambiguous. Negative examples highlight typical sources of rejection:  (1) conflicting words, e.g., the change is ''cake → table`` but the image contains both a table and a cake; (2) problematic questions, e.g., the change is ''baby → dog`` but the question implies an animal; and (3) problematic answers, e.g., the color of the field could be described as being ''red`` (distractor answer).}
    \label{fig:human_validation_examples}
\end{figure*}

\begin{figure*}
    \centering
    \includegraphics[width=1\linewidth]{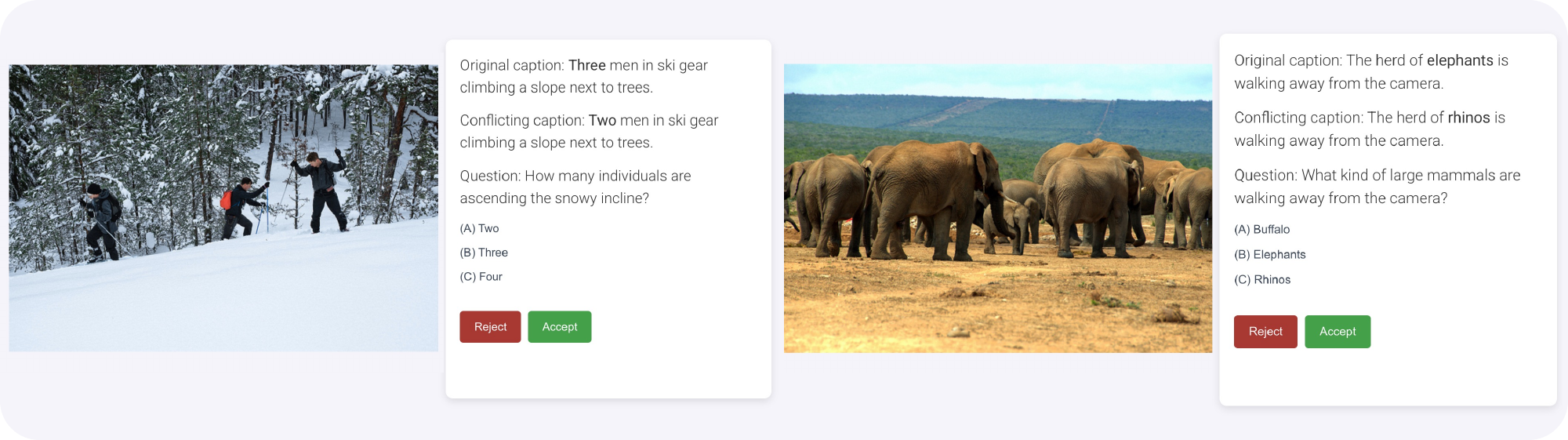}
    \caption{Examples from the human verification interface. Each sample includes an original caption from MS COCO, a conflicting caption that introduces a controlled contradiction, a targeted question designed to test conflict detection, and multiple-choice answers. Annotators use binary accept/reject voting to validate the quality and clarity of each sample, ensuring the reliability of the benchmark’s diagnostic test set.}
    \label{fig:examples_from_website}
\end{figure*}

\section{Broader impact}
\label{app:sec:broader_impact}
This work contributes to the development of more reliable multimodal AI systems by exposing critical limitations in conflict detection capabilities. Improved conflict detection could enhance AI safety in applications like medical diagnosis, autonomous systems, and content verification. However, the focus on synthetic contradictions may not fully represent the complexity of real-world misinformation or adversarial scenarios. We encourage future work to extend these findings to more diverse contradiction types and real-world deployment contexts.

\end{document}